\documentclass[runningheads]{llncs}

\usepackage{graphicx}

\usepackage{tikz}
\usepackage{comment}
\usepackage{amsmath,amssymb} 
\usepackage{color}
\usepackage[accsupp]{axessibility}  
\usepackage{hyperref}
\usepackage{kotex} 
\usepackage{booktabs}
\usepackage[ruled,vlined]{algorithm2e}
\usepackage{amsmath}
\usepackage{wrapfig}
\definecolor{commentcolor}{RGB}{110,154,155}   

\usepackage{subcaption}

\begin{document}
\pagestyle{headings}
\mainmatter
\def\ECCVSubNumber{6403}  

\title{High-Resolution Virtual Try-On with \\ Misalignment and Occlusion-Handled Conditions}

\titlerunning{HR-VITON with Misalignment and Occlusion-Handled Conditions}
%
\author{Sangyun Lee\inst{1,*} \and
Gyojung Gu\inst{2,3,*} \and \\
Sunghyun Park\inst{2} \and
Seunghwan Choi\inst{2} \and
Jaegul Choo\inst{2}}

\authorrunning{Lee et al.}
%
\institute{Soongsil University \and Korea Advanced Institute of Science and Technology \and Nestyle Inc.\\
\email{ml.swlee@gmail.com \tt \{gyojung.gu, psh01087, shadow2496, jchoo\}@kaist.ac.kr} \\
* indicates equal contributions.}
\maketitle

\begin{center}
    \centering 
    \includegraphics[width=\linewidth]{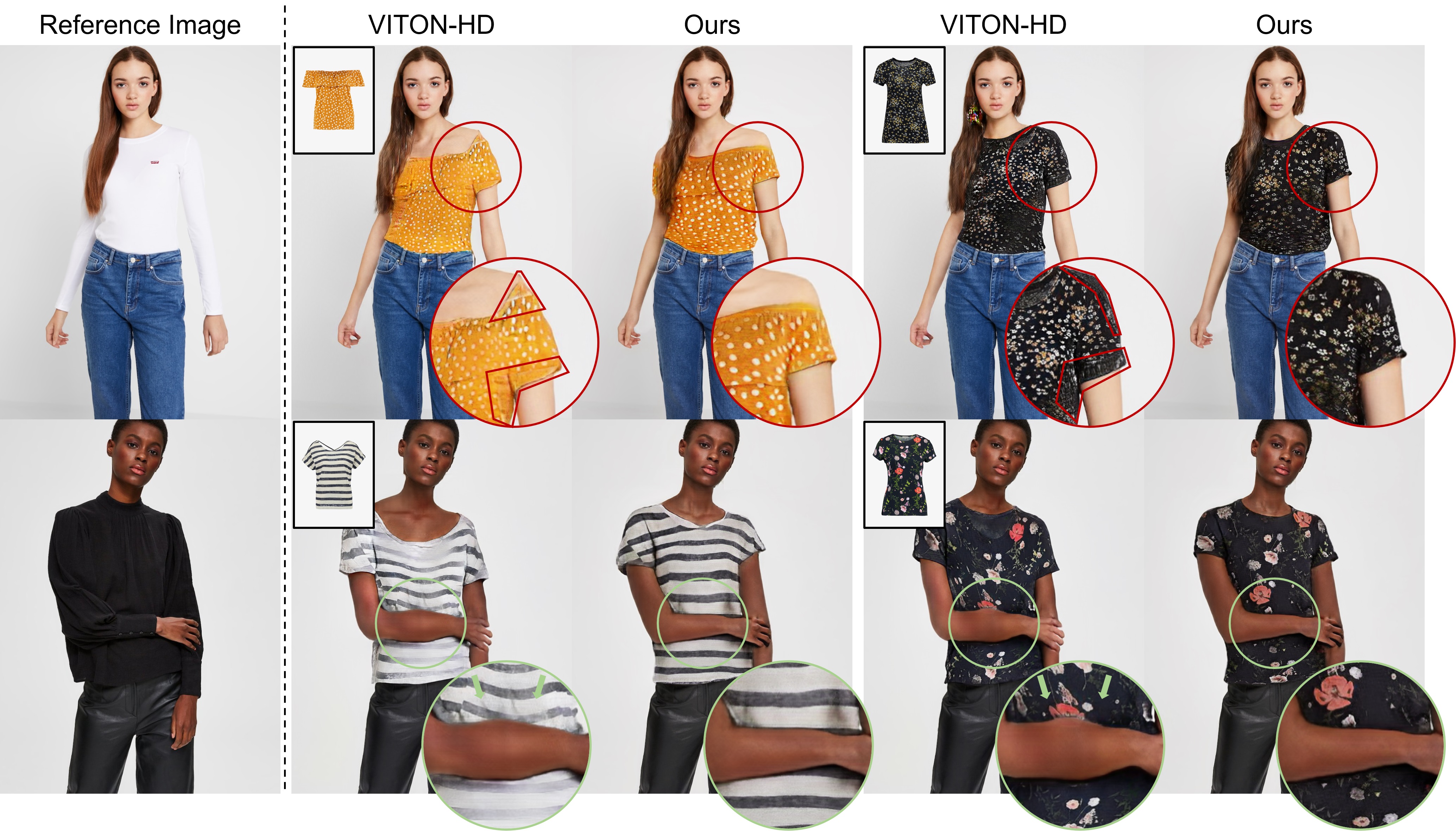}
    \captionof{figure}{Comparison of 1024$\times$768 try-on synthesis results with VITON-HD~\cite{choi2021viton}. 
    ($1st$ row) The red-colored areas indicate the artifact due to the misalignment between a warped clothing image and a segmentation map.
    ($2nd$ row) The green-colored areas denote the pixel-squeezing due to the occlusion by the body parts.
    In contrast to the VITON-HD, our method successfully handles the misalignment and occlusion.
    Zoom in for the best view.
    }
    \label{fig:main}
\end{center}


\begin{abstract}

Image-based virtual try-on aims to synthesize an image of a person wearing a given clothing item.
To solve the task, the existing methods warp the clothing item to fit the person's body and generate the segmentation map of the person wearing the item before fusing the item with the person.
However, when the warping and the segmentation generation stages operate individually without information exchange, the misalignment between the warped clothes and the segmentation map occurs, which leads to the artifacts in the final image.
The information disconnection also causes excessive warping near the clothing regions occluded by the body parts, so-called pixel-squeezing artifacts.
To settle the issues, we propose a novel try-on condition generator as a unified module of the two stages (\textit{i.e.}, warping and segmentation generation stages).
A newly proposed feature fusion block in the condition generator implements the information exchange, and the condition generator does not create any misalignment or pixel-squeezing artifacts. 
We also introduce discriminator rejection that filters out the incorrect segmentation map predictions and assures the performance of virtual try-on frameworks.
Experiments on a high-resolution dataset demonstrate that our model successfully handles the misalignment and occlusion, and significantly outperforms the baselines.
Code is available at \href{https://github.com/sangyun884/HR-VITON}{https://github.com/sangyun884/HR-VITON}.

\keywords{High-Resolution Virtual Try-On, Misalignment-Free, Occlusion-Handling}
\end{abstract}

\section{Introduction}

As the importance of online shopping increases, a technology that allows customers to virtually try on clothes is expected to enrich the customer's experience.
A virtual try-on task aims to change the clothing item on a person into a given clothing product.
While there are 3D-based virtual try-on approaches that rely on the 3D measurement of garments~\cite{guan2012drape,sekine2014virtual,pons2017clothcap,patel2020tailornet}, we address image-based virtual try-on~\cite{jetchev2017conditional,han2018viton,wang2018toward,yu2019vtnfp,yang2020towards,chopra2021zflow,li2021toward,lewis2021vogue}, which only requires a garment and a person image, facilitating real-world applications.

To address this task, previous studies employ an explicit warping module that aligns the clothing image with the person's body.
Moreover, predicting the segmentation map of the final image alleviates the difficulty of image generation as it guides the person's layout and separates regions to be generated and the ones to be preserved~\cite{yang2020towards}.
The importance of the segmentation map increases as the image resolution grows.
Most image-based virtual try-on methods include these stages~\cite{han2018viton,wang2018toward,yu2019vtnfp,yang2020towards,choi2021viton,chopra2021zflow,li2021toward}, and the outputs of the warping and segmentation map generation modules greatly influence the final try-on results.

However, the virtual try-on frameworks that consist of warping and segmentation generation modules have misaligned regions between the warped clothes and the segmentation map, so-called \emph{misalignment}.
As shown in Fig.~\ref{fig:main}, the misalignment results in the artifacts in these regions, which harm the perceptual quality of the final result significantly, especially at the high resolution.
The main cause of misalignment is that the warping module and the segmentation map generator operate separately without information exchange.
Although a recent study~\cite{choi2021viton} tries to alleviate the artifacts in the misaligned regions, the existing methods are still not possible to solve the misalignment problem completely.

The information disconnection between two modules yields another problem (\textit{i.e.}, pixel-squeezing artifacts).
As shown in Fig.~\ref{fig:main}, the results of the previous methods are significantly impaired when the body parts occlude the garment.
Pixel-squeezing artifacts are caused by excessive warping of clothes near the occluded regions, which is due to the lack of information exchange between the warping and the segmentation map generation modules.
The artifacts limit the possible poses of the person images, making it difficult to apply virtual try-on to the real world.

To settle the issues, we propose a novel try-on condition generator that unifies the warping and segmentation generation modules.
The proposed module simultaneously predicts the warped garment and the segmentation map, which are perfectly aligned to each other.
Our try-on condition generator can remove the misalignment completely and handle the occlusions by the body parts naturally.
Extensive experiments show that the proposed framework successfully handles the occlusion and misalignment, and achieves state-of-the-art results on the high-resolution dataset (\textit{i.e.}, 1024$\times$768), both quantitatively and qualitatively.

In addition, we introduce a discriminator rejection that filters out incorrect segmentation map predictions, which lead to unnatural final results.
We demonstrate that the discriminator rejection assures the performance of virtual try-on frameworks, which is an important feature for real-world applications.

We summarize our contributions as follows:
\begin{itemize}
    \item We propose a novel architecture that performs warping and segmentation map generation simultaneously.
    \item Our method is inherently \emph{misalignment-free} and can handle the occlusion of clothes by body parts naturally.
    \item We adapt the discriminator rejection to filter out incorrect segmentation map predictions. 
    \item We achieve state-of-the-art performance on a high-resolution dataset.
\end{itemize}

\section{Related Work}

\subsection{Image-based Virtual Try-On}
An image-based virtual try-on task aims to produce a person image wearing a target clothing item given a pair of clothes and person images.
Recent virtual try-on methods~\cite{han2018viton,wang2018toward,yu2019vtnfp,yang2020towards,choi2021viton,chopra2021zflow,li2021toward} generally consist of three separate modules: 1) segmentation map generation module, 2) clothing warping module, and 3) fusion module.
The fusion module can generate the photo-realistic images by utilizing intermediate representations such as warped clothes and segmentation maps, which are produced by previous stages.

\noindent\textbf{Clothes Deformation.}
To preserve the details of a clothing item, previous approaches~\cite{han2018viton,wang2018toward,han2019clothflow,chopra2021zflow} rely on the explicit warping module to fit the input clothing item to a given person's body.
VITON~\cite{han2018viton} and CP-VTON\cite{wang2018toward} predict the parameters for thin plate spline (TPS) transformation to warp the clothing item.
Since the warping modules based on the TPS transformation have a limited degree of freedom, an appearance flow is utilized to compute a pixel-wise 2D deformation field of the clothing image~\cite{han2019clothflow,chopra2021zflow}.
Although the warping modules have been consistently improved, the misalignment between the warped clothes and a person's body remains and results in the artifacts in the misaligned regions.
Recently, VITON-HD~\cite{choi2021viton} proposed a normalization technique to alleviate the issue.
However, we found that the normalization method fails to naturally fill the misaligned regions with clothing texture.
In this paper, we propose a method that can generate warped clothes without misaligned regions.

\noindent\textbf{Segmentation Generation for Try-On Synthesis.}
To guide the try-on image synthesis, recent virtual try-on models~\cite{jandial2020sievenet,yang2020towards,minar2020cloth,xie2021vton,chopra2021zflow,li2021toward} utilize the human segmentation maps of a person wearing the target clothes.
The segmentation map disentangles the generation of appearance and shape, allowing the model to produce more spatially coherent results.
In particular, the high-resolution virtual try-on methods~\cite{choi2021viton,li2021toward} generally include the segmentation generation module because the importance of the segmentation map increases as the image resolution grows.

\subsection{Rejection Sampling}
There are several studies that aim to reject the low-quality generator outputs to improve the fidelity of samples.
Razavi \textit{et al.}~\cite{razavi2019generating} introduced rejection sampling based on the probability that the pre-trained classifier assigns to the correct class. 
Azadi \textit{et al.}~\cite{azadi2018discriminator} proposed the discriminator rejection sampling, where a discriminator rejects the generated samples at test time. 
Under strict assumptions, this allows exact sampling from the data distribution. 
Although there have been several follow-up works~\cite{turner2019metropolis,mo2019mining}, this technique has not been commonly used for image-conditional generation. 
In this paper, we utilize the discriminator to filter out the low-quality samples at test time.

\section{Proposed Method}

Given a reference image $I \in \mathbb{R}^{3 \times H \times W}$ of a person and a clothing image $c \in \mathbb{R}^{3 \times H \times W}$ ($H$ and $W$ denote the image height and width, respectively), our goal is to synthesize an image $\hat{I} \in \mathbb{R}^{3 \times H \times W}$ of the person wearing $c$, where the pose and the body shape of $I$ are maintained.
Following the training procedure of VITON~\cite{han2018viton}, we train the model to reconstruct $I$ from a clothing-agnostic person representation and $c$ that the person is wearing already.
The clothing-agnostic person representation eliminates any clothing information in $I$, and it allows the model to generalize at test time when an arbitrary clothing image is given.

\begin{figure}[t!]
    \centering
    \includegraphics[width=1.0\linewidth]{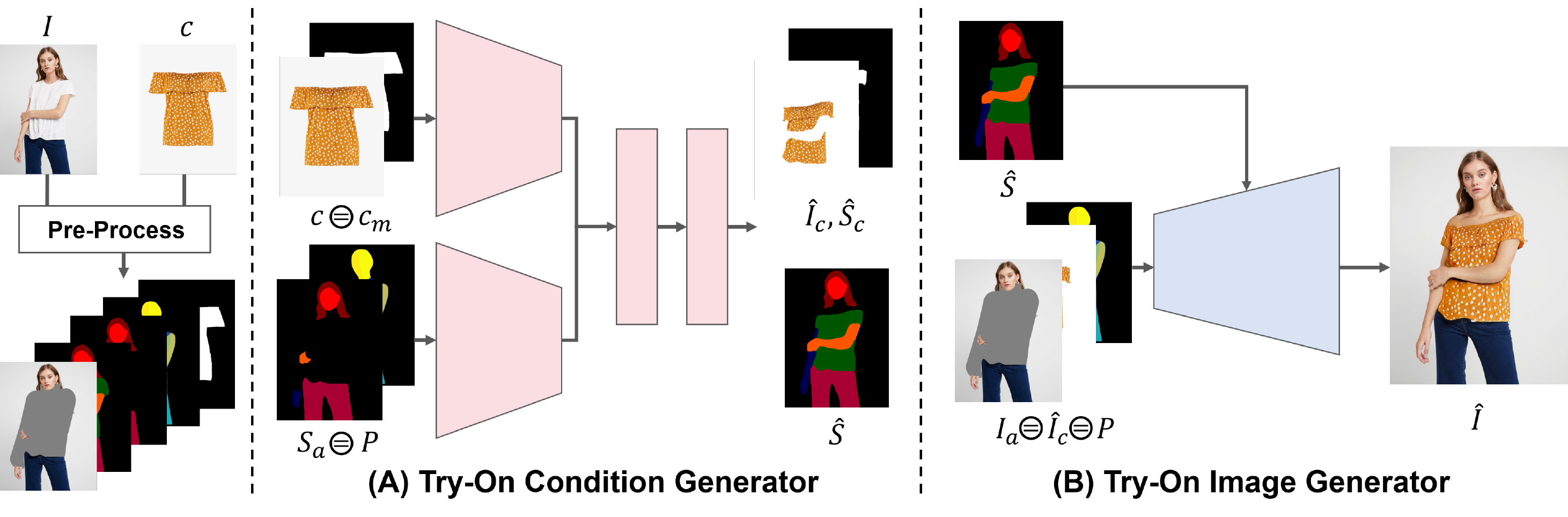}
    \caption{Overview of the proposed framework (HR-VITON).}
    \label{fig:overview}
\end{figure}

Our framework is composed of two stages: (1) a \emph{try-on condition generator}; (2) a \emph{try-on image generator} (see Fig.~\ref{fig:overview}).
Given the clothing-agnostic person representation and $c$, our try-on condition generator deforms $c$ and produces the segmentation map simultaneously.
The generator does not create any misalignment or pixel-squeezing artifacts (Section~\ref{sec:TOCG}).
Afterward, the try-on image generator synthesizes the final try-on result using the outputs of the try-on condition generator (Section~\ref{sec:TOIG}).
At test time, we apply discriminator rejection that filters out incorrect segmentation map predictions (Section~\ref{sec:rejection}).

\noindent\textbf{Pre-Processing.}
In the pre-processing step, we obtain a segmentation map $S \in \mathbb{L}^{H \times W}$ of the person, a clothing mask $c_m \in \mathbb{L}^{H \times W}$, and a pose map $P \in \mathbb{R}^{3 \times H \times W}$ with the off-the-shelf models~\cite{gong2018instance,guler2018densepose}, where $\mathbb{L}$ is a set of integers indicating the semantic labels.
For the pose map $P$, we utilize a dense pose~\cite{guler2018densepose}, which maps all pixels of the person regions in the RGB image to the 3D surface of the person's body.
For the clothing-agnostic person representation, we employ a clothing-agnostic person image $I_a$ and a clothing-agnostic segmentation map $S_a$ as those of VITON-HD~\cite{choi2021viton}.

\subsection{Try-On Condition Generator}

\label{sec:TOCG}

\begin{figure}[t!]
    \centering
    \includegraphics[width=1.0\linewidth]{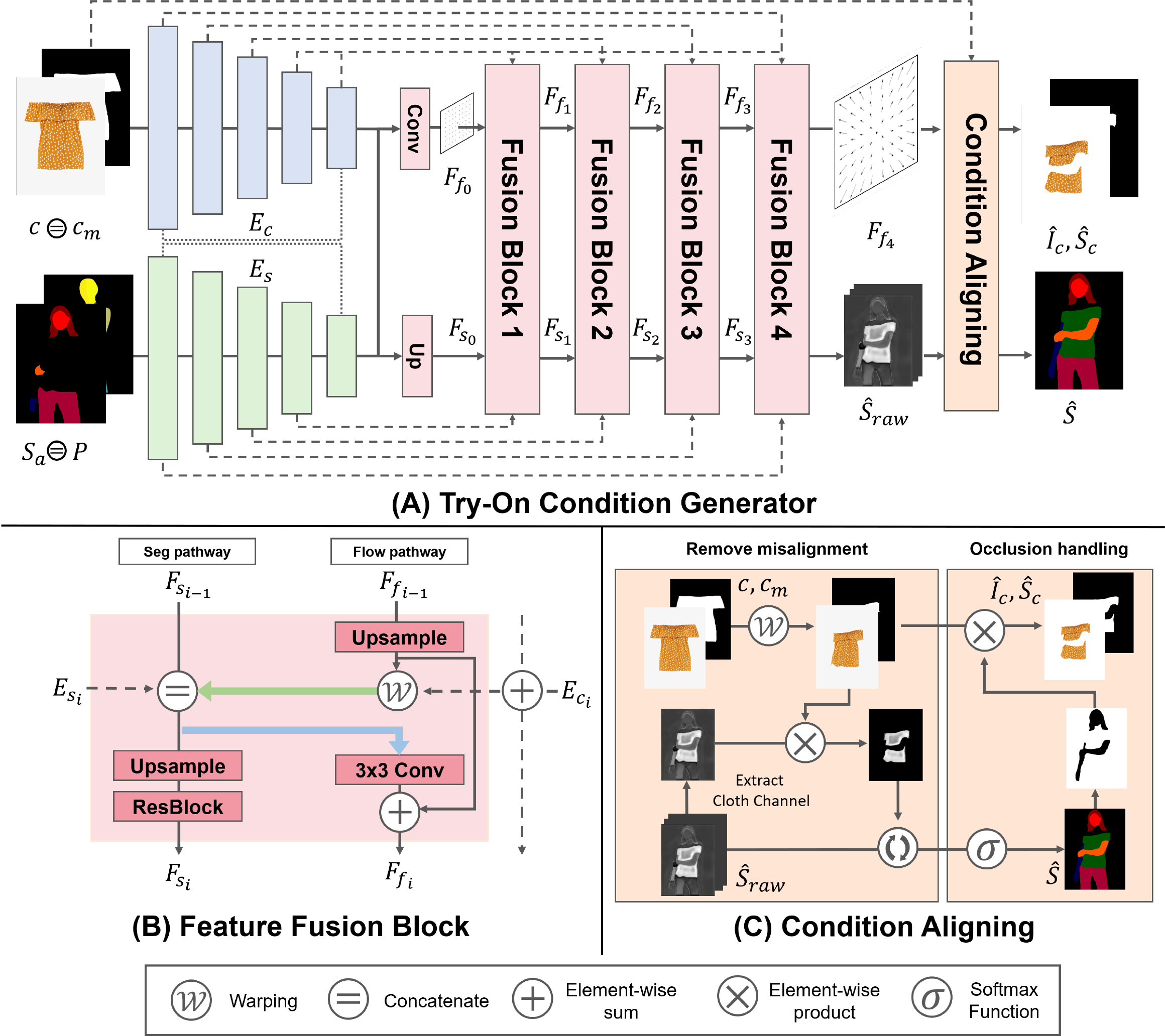}
    \caption{Architecture of try-on condition generator.}
    \label{fig:condition_generator}
\end{figure}

In this stage, we aim to generate the segmentation map $\hat{S}$ of the person wearing the target clothing item $c$ and deform $c$ to fit the body of the person.
A warped clothing image $\hat{I}_c$ and a generated segmentation map $\hat{S}$ are used as the conditions for the try-on image generator.
Fig.~\ref{fig:condition_generator} (A) shows the overall architecture of our try-on condition generator.
Our try-on condition generator consists of two encoders (\textit{i.e.}, a clothing encoder $E_{c}$ and a segmentation encoder $E_{s}$) and a decoder.
Given $(c, c_m)$ and $(S_a, P)$, we first extract the feature pyramid $\{E_{c_k}\}_{k=0}^4$ and $\{E_{s_l}\}_{l=0}^4$ from each encoder, respectively.
The extracted features are fed into the feature fusion blocks of the decoder, where the feature maps obtained from the two different feature pyramids are fused to predict the segmentation map and the appearance flow for warping the clothing image.
Given the outputs of the last feature fusion block, we obtain $\hat{I}_c, \hat{S}_c$, and $\hat{S}$ through condition aligning.

\noindent\textbf{Feature Fusion Block.}
As shown in Fig. ~\ref{fig:condition_generator} (B), there are two pathways in the feature fusion block: the \emph{flow pathway} and the \emph{seg pathway}.
The flow and seg pathway generate the appearance flow map $F_{f_{i}}$ and the segmentation feature $F_{s_{i}}$, respectively. 
These two pathways exchange information with each other to estimate the appearance flow and the segmentation map jointly, which is indicated by green and blue arrows.
For the green arrow, $F_{f_{i-1}}$ is used to deform the feature extracted from $c$ and $c_m$, which is then concatenated with $F_{s_{i-1}}$ and $E_{s_{i}}$ to generate $F_{s_{i}}$. For the blue arrow, $F_{s_{i-1}}$ is used to guide the flow estimation.
These information exchanges are crucial in estimating the warped clothing and the segmentation map aligned each other.
The feature fusion block estimates $F_{f_i}$ and $F_{s_i}$ simultaneously, which are then used to refine each other at the next block.

\noindent\textbf{Condition Aligning.}
To prevent the misalignment, we obtain \(\hat S\) by removing the non-overlapping regions of the clothing mask channel of \(\hat S_{raw}^{k,i,j}\) with \(W(c_m, F_{f_4})\):
\begin{equation}
    \hat S_{logit}^{k,i,j} =
    \begin{cases}
        \hat S_{raw}^{k,i,j} &\text{if\ \ $k\neq C$}\\
        \hat S_{raw}^{k,i,j} \cdot W(c_m, F_{f_4}) &\text{if\ \ $k=C$}
    \end{cases}
\end{equation}
\begin{equation}
    \hat S = \sigma(\hat S_{logit}),
\end{equation}
where $\hat S_{raw}$ is equivalent to $F_{s_4}$ and \(C\) denotes the index of the clothing mask channel.
\(i,j\), and \(k\) are indices across the spatial and channel dimensions.
\(\sigma\) is depth-wise softmax. Note that we apply ReLU activation to assure that $\hat S_{raw}$ is nonnegative.

\(\hat I_c\) and \(\hat S_c\) are obtained by applying the body part occlusion handling to \(W(c, F_{f_4})\).
As Fig.~\ref{fig:condition_generator} (C) demonstrates, the body parts of \(\hat S\) are used to remove the occluded regions from \(W(c,F_{f_4})\) and \(W(c_m,F_{f_4})\).
Body part occlusion handling helps to eliminate the pixel-squeezing artifacts (see Fig.~\ref{fig:occlusion}).

\noindent\textbf{Loss Functions.}
We use the pixel-wise cross-entropy loss $\mathcal{L}_{CE}$ between predicted segmentation map $\hat S$ and $S$.
Additionally, $L1$ loss and perceptual loss are used to encourage the network to warp the clothes to fit the person's pose. These loss functions are also directly applied to the intermediate flow estimations to prevent the intermediate flow maps from vanishing and improve the performance.
Formally, $\mathcal{L}_{L1}$ and $\mathcal{L}_{VGG}$ are as follows:
\begin{equation}
    \label{eqn:l1}
    \mathcal{L}_{L1} = \sum_{i=0}^{3}w_i\cdot ||W(c_m, F_{f_i}) - S_c||_1 + ||\hat{S}_c - S_c||_1,
\end{equation}
\begin{equation}
    \label{eqn:vgg}
     \mathcal{L}_{VGG} = \sum_{i=0}^{3}w_i\cdot \phi(W(c, F_{f_i}), I_c) + \phi(\hat{I}_c, I_c),
\end{equation}
where $w_i$ determines the relative importance between each terms.

$\mathcal{L}_{TV}$ is a total-variation loss to enforce the smoothness of the appearance flow:
\begin{equation}
    \mathcal{L}_{TV} = ||\nabla F_{f_4}||_1
\end{equation}
We found that regularizing only the last appearance flow $F_{f_4}$ is vital in learning the flow estimation at coarse scales.

Totally, our try-on condition generator is trained end-to-end using the following objective function:
\begin{equation}
    \mathcal{L}_{TOCG} = \lambda_{CE}\mathcal{L}_{CE} + \mathcal{L}_{cGAN} + \lambda_{L1}\mathcal{L}_{L1} + \mathcal{L}_{VGG} + \lambda_{TV}\mathcal{L}_{TV},
\end{equation}
where \(\mathcal{L}_{cGAN}\) is conditional GAN loss between $\hat S$ and $S$, and $\lambda_{CE}$, $\lambda_{L1}$, and $\lambda_{TV}$ denote the hyper-parameters controlling relative importance between different losses.
For \(\mathcal{L}_{cGAN}\), we used the least-squared GAN loss~\cite{mao2017least}.

\subsection{Try-On Image Generator}
\label{sec:TOIG}
In this stage, we generate the final try-on image $\hat{I}$ by fusing the clothing-agnostic image $I_a$, the warped clothing image $\hat{I}_c$, and the pose map $P$, guided by $\hat{S}$.
The try-on image generator consists of a series of residual blocks, along with upsampling layers.
The residual blocks use SPADE~\cite{park2019semantic} as normalization layers whose modulation parameters are inferred from $\hat{S}$.
Also, the input $(I_a, \hat{I}_c, P)$ is resized and concatenated to the activation before each residual block.
We train the generator with the same losses used in SPADE and pix2pixHD~\cite{wang2018high}.
Details of the model architecture, hyperparameters, and the objective function are described in the appendix.

\subsection{Discriminator Rejection}
\label{sec:rejection}
We propose a discriminator rejection method to filter out the low-quality segmentation map generated by the try-on condition generator at the test time.
In the discriminator rejection sampling~\cite{azadi2018discriminator}, the acceptance probability for an input \(x\) is
\begin{equation}
    p_{accept}(x) = \frac{p_d(x)}{Lp_g(x)},
\end{equation}
where \(p_d\) and \(p_g\) are the data distribution and the implicit distribution given by the generator, and \(L\) is a normalizing constant. As we use the least-squares GAN loss, the optimal discriminator is derived as follows:
\begin{equation}
    D^*(x) = \frac{p_d(x)}{p_d(x)+p_g(x)}
\end{equation}

Afterward, the acceptance probability can be represented using the discriminator $D(x)$:
\begin{equation}
    p_{accept}=\frac{D(x)}{L(1-D(x))},
\end{equation}
Where the equality is satisfied only if \(D=D^*\). \(L\) is written as follows:
\begin{equation}
    L=\max_x\frac{D(x)}{(1-D(x))},
\end{equation}
which is intractable. In practice, we construct \(x\) from the segmentation map and input conditions (i.e., $P, S_a, c,$ and $c_m$) and obtain \(L\) using the entire training dataset.
Azadi \textit{et al.}~\cite{azadi2018discriminator} sample \(\psi\sim U(0,1)\) and reject \(x\) if \(\psi > p_{accept}(x)\).
Instead, we reject \(x\) if \(p_{accept}(x)\) is below a certain threshold. The discriminator rejection enables us to filter out the incorrect segmentation maps faithfully.

\begin{figure}[t!]
    \centering
    \includegraphics[width=1.0\linewidth]{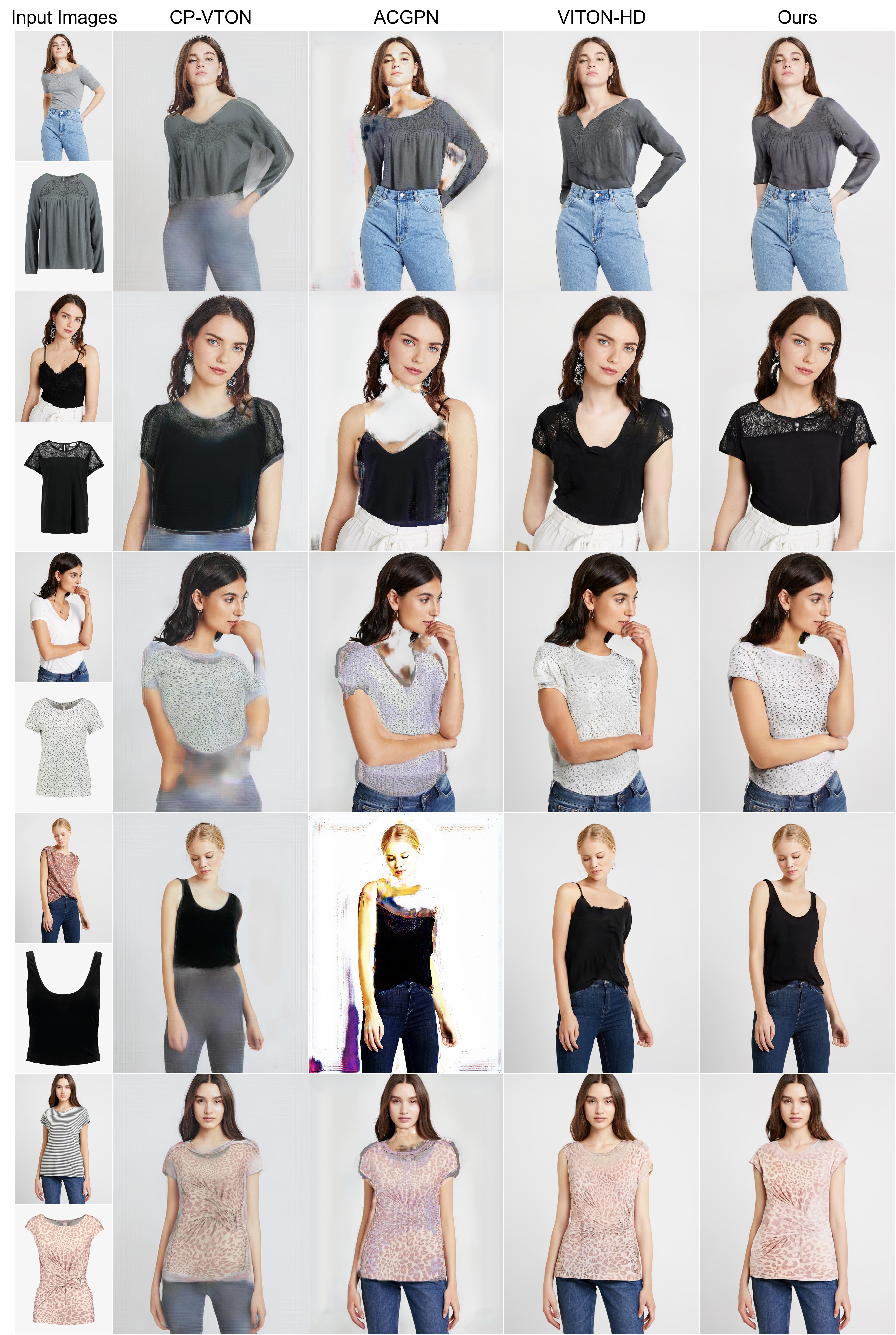}
    \vspace{-0.5cm}
    \caption{Qualitative comparison with baselines.}
    \vspace{-0.5cm}
    \label{fig:viscomp_img}
\end{figure}

\section{Experiments}

\subsection{Training}

For the experiments, we use a high-resolution virtual try-on dataset introduced by VITON-HD~\cite{choi2021viton}, which contains 13,679 frontal-view woman and top clothing image pairs.
The original resolution of the images is 1024$\times$768, and the images are bicubically downsampled to the desired resolutions when needed.
We split the dataset into a training and a test set with 11,647 and 2,032 pairs, respectively.
For detailed information on the model training, see appendix.

\subsection{Qualitative Results}

\begin{figure}[t!]
    \centering
    \includegraphics[width=1.0\linewidth]{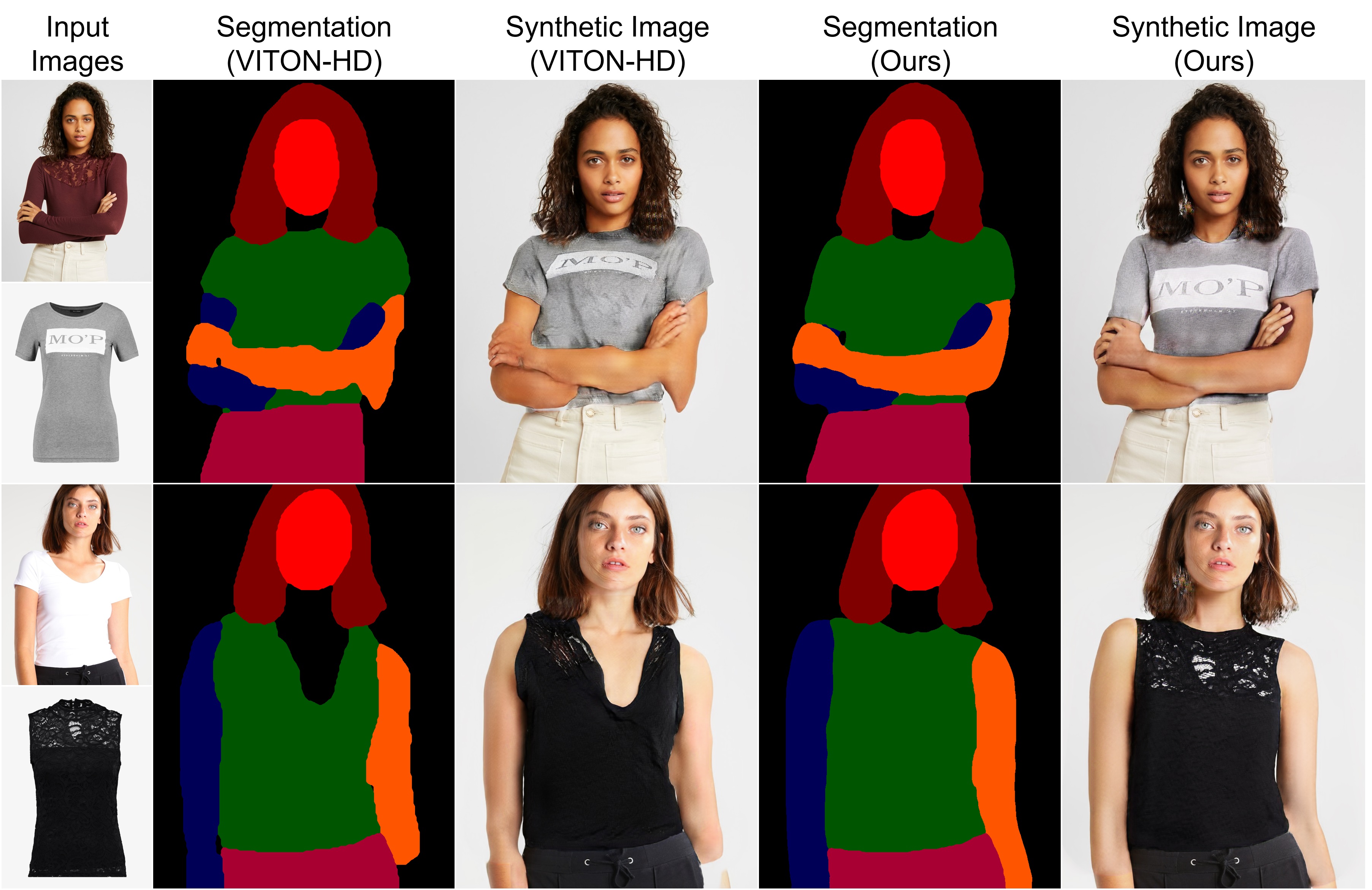}
    \caption{Try-on synthesis results and corresponding segmentation maps.}
    \label{fig:viscomp_seg}
\end{figure}

\begin{figure}[t!]
    \centering
    \includegraphics[width=1.0\linewidth]{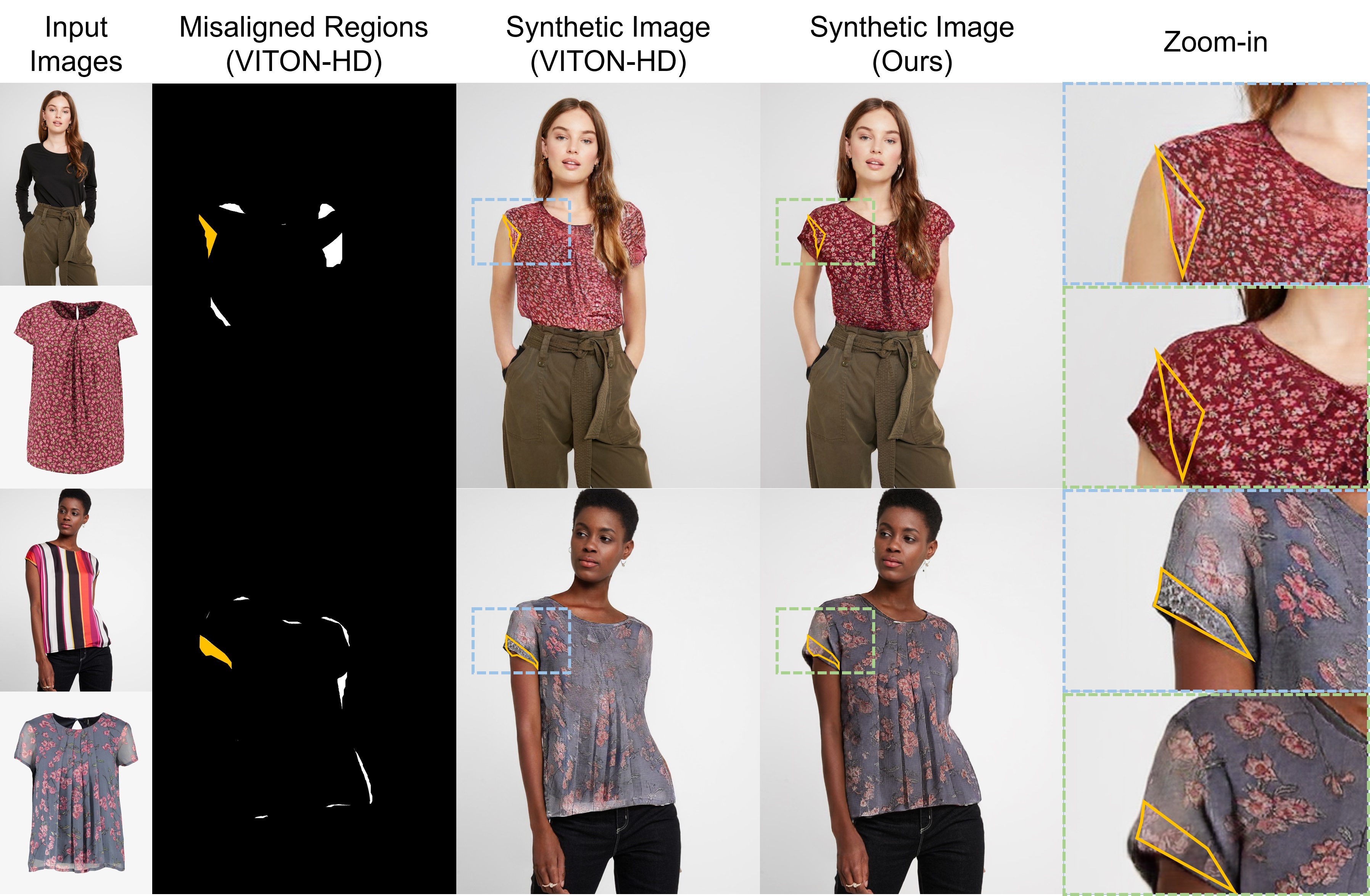}
    \caption{Synthesis results and corresponding misaligned regions indicated by yellow colored areas. VITON-HD suffers from the artifacts caused by misalignment.}
    \label{fig:misalignment}
    \vspace{0.5cm}
\end{figure}

\noindent\textbf{Comparison with Baselines.}
We compare our method with several state-of-the-art baselines, including CP-VTON~\cite{wang2018toward}, ACGPN~\cite{yang2020towards}, and VITON-HD~\cite{choi2021viton}.
We utilize the publicly available codes for baselines.
Fig.~\ref{fig:viscomp_img} shows that our method generates more photo-realistic images compared to the baselines.
Specifically, we observe that our model not only preserves the details of the target clothing images but also generates the neckline naturally.
As shown in Fig.~\ref{fig:viscomp_seg}, our try-on condition generator has the capability to produce the body shape more naturally compared to VITON-HD.
These results demonstrate that the quality of the conditions for the try-on image generator is crucial in achieving perceptually convincing results.
Furthermore, Fig.~\ref{fig:misalignment} shows that VITON-HD fails to eliminate the artifacts in the misaligned regions completely.
On the other hand, since our method can produce misalignment-free segmentation maps and warped clothing images, our method solves the misalignment problem inherently.
Thus, our method successfully synthesizes the high-quality images.

\begin{figure}[t!]
    \centering
    \includegraphics[width=1.0\linewidth]{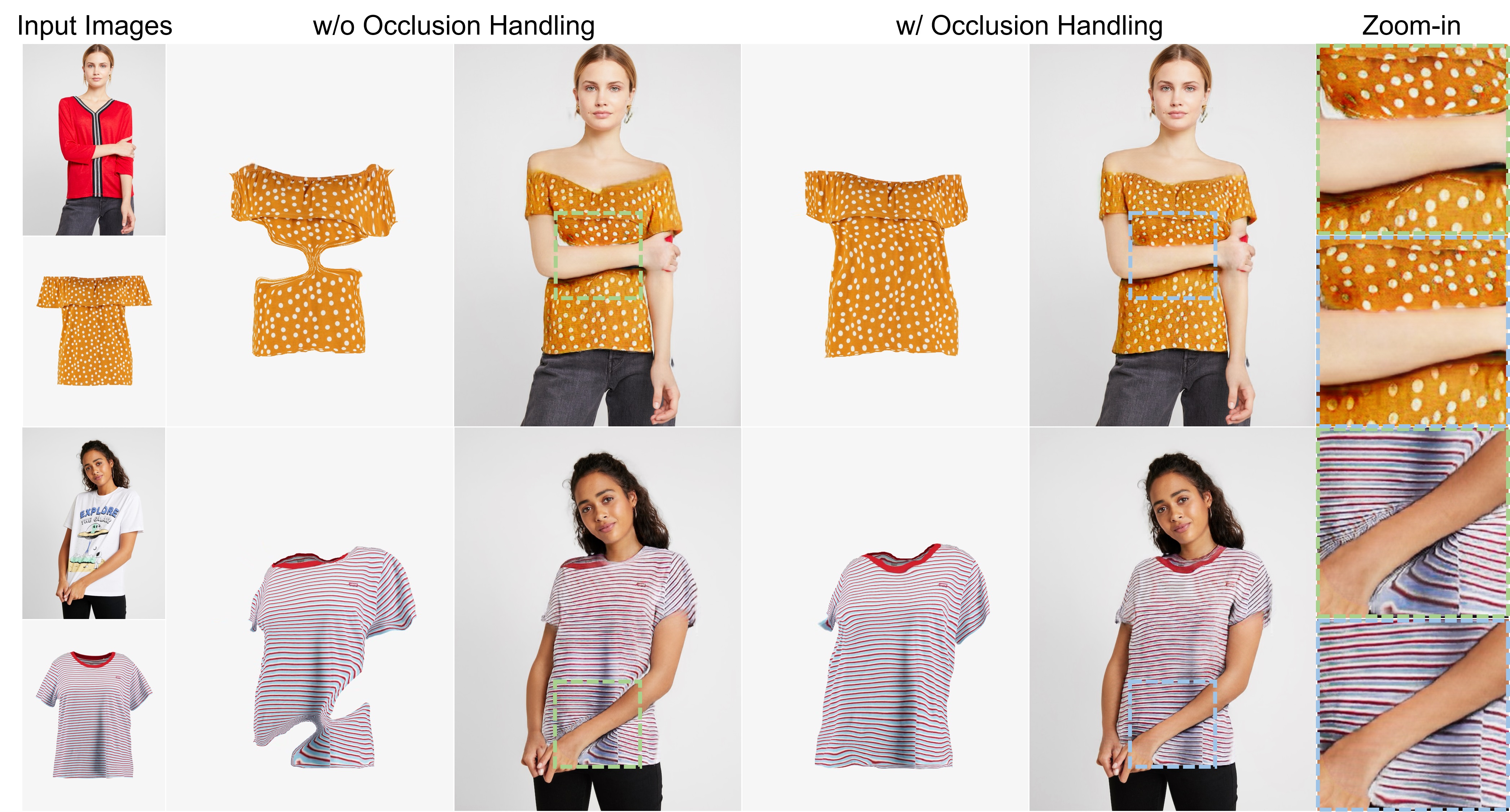}
    \caption{Effects of the body part occlusion handling. The green colored areas indicate the pixel-squeezing artifacts.}
    \label{fig:occlusion}
    \vspace{0.5cm}
\end{figure}

\noindent\textbf{Effectiveness of Occlusion Handling.}
We analyze the impact of the occlusion handling process in our try-on condition generator.
Fig.~\ref{fig:occlusion} shows the effectiveness of the proposed body part occlusion handling. 
Without occlusion handling, the model excessively deforms the clothing image to fit the person's body shape, as shown in the $2nd$ column of Fig.~\ref{fig:occlusion}.
Due to the undesired deformation, the texture (\textit{e.g.}, logo and stripe) of the target clothing item is squeezed, causing the missing pattern in the final results (See the $3rd$ column of Fig.~\ref{fig:occlusion}).
On the other hand, the model with occlusion handling enables to warp the clothes without the pixel-squeezing, better preserving the high-frequency details of the garment.

\begin{figure}[t!]
    \includegraphics[width=1.0\linewidth]{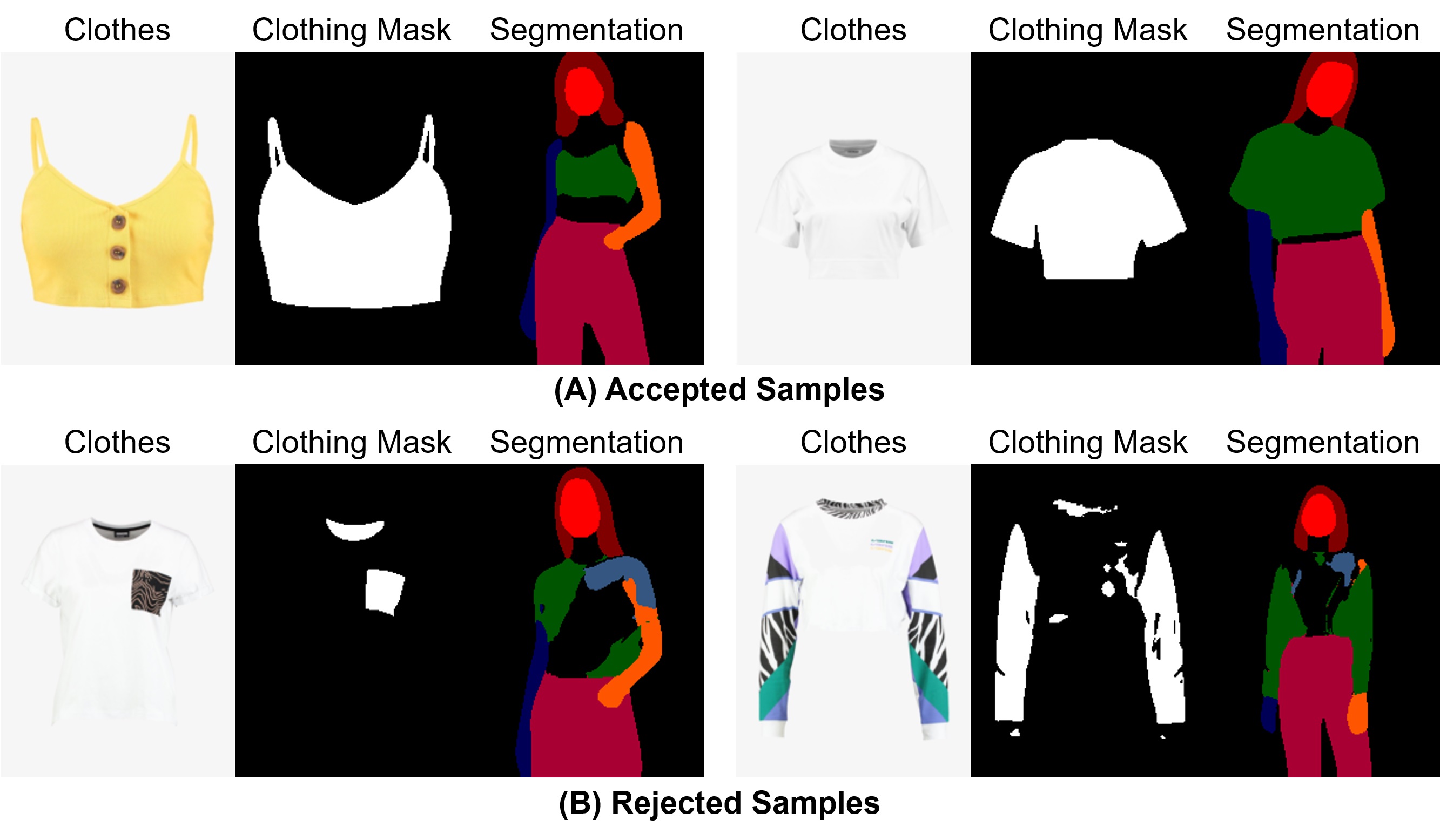}
    \caption{Examples of accepted (A) and rejected (B) segmentation maps by discriminator rejection, corresponding input clothes and clothing masks.}
    \label{fig:rejection}
\end{figure}

\noindent\textbf{Effectiveness of Discriminator Rejection.}
To filter out the low-quality segmentation maps produced by our try-on condition generator, we propose a discriminator rejection method.
Fig.~\ref{fig:rejection} shows the accepted and the rejected samples of our discriminator rejection.
Different from the accepted samples, the segmentation maps of the rejected samples are considerably impaired, as shown in the $2nd$ row of Fig.~\ref{fig:rejection}.
We found that the incorrect segmentation maps are caused mainly by errors in the pre-processing step, such as obtaining the clothing mask.
Most virtual try-on methods rely on multiple conditions such as segmentation map and pose information obtained in the pre-processing stage and thus are prone to these errors.
We believe that our discriminator rejection method can be a simple and effective solution for filtering out the low-quality outputs.

\subsection{Quantitative Results}
\begin{table}[b!]
    \centering
    \footnotesize
    \begin{tabular}{l|cc}
    \toprule
    Method & FID$_{\downarrow}$ & KID$_{\downarrow}$  \\ 
    \cmidrule(lr){1-3}
    HR-VITON & \textbf{10.91} & \textbf{0.179} \\ 
    $\llcorner$ w/o Condition Aligning                     & 12.05 & 0.356 \\
    $\llcorner$ w/o Feature Fusion Block                        & 12.41 & 0.381 \\
    $\llcorner$ w/o Feature Fusion Block \& Condition Aligning*     & 12.73 & 0.415 \\
    \bottomrule
    \end{tabular}
    \caption{Ablation study in unpaired setting. We describes the KID as a value multiplied by 100. *Last row denotes that there is no information exchange.}
    
    \label{table:ablation}
\end{table}

\begin{table}[]
\centering
\scriptsize
\begin{tabular}{l|cccc|cccc|cccc}
    \toprule
     & \multicolumn{4}{c}{256$\times$192} & \multicolumn{4}{c}{512$\times$384} & \multicolumn{4}{c}{1024$\times$768}\\ 
     & LPIPS$_{\downarrow}$ & SSIM$_{\uparrow}$ & FID$_{\downarrow}$ & KID$_{\downarrow}$ & LPIPS$_{\downarrow}$ & SSIM$_{\uparrow}$ & FID$_{\downarrow}$ & KID$_{\downarrow}$ & LPIPS$_{\downarrow}$ & SSIM$_{\uparrow}$ & FID$_{\downarrow}$ & KID$_{\downarrow}$ \\ 
    \midrule
    CP-VTON  & 0.159 & 0.739 & 30.11 & 2.034 
             & 0.141 & 0.791 & 30.25 & 4.012
             & 0.158 & 0.786 & 43.28 & 3.762
             \\
    ACGPN    & 0.074 & 0.833 & 11.33 & 0.344
             & 0.076 & 0.858 & 14.43 & 0.587
             & 0.112 & 0.850 & 43.29 & 3.730
             \\
    VITON-HD & 0.084 & 0.811 & 16.36 & 0.871
             & 0.076 & 0.843 & 11.64 & 0.300
             & 0.077 & 0.873 & 11.59 & 0.247
             \\
    PF-AFN & - & - & - & -
             & - & - & - &
             & - & - & 14.01 & 0.588
             \\
    \midrule
    HR-VITON    & \textbf{0.062} & \textbf{0.864} & \textbf{9.38} & \textbf{0.153}   
             & \textbf{0.061} & \textbf{0.878} & \textbf{9.90} & \textbf{0.188}
             & \textbf{0.065} & \textbf{0.892} & \textbf{10.91} & \textbf{0.179}
             \\ 
    \bottomrule
\end{tabular}
\caption{Quantitative comparison with baselines. 
         We describes the KID as a value multiplied by 100. HR-VITON refers to our model.}
\label{table:main}
\end{table}

\begin{figure}[b!]
    \includegraphics[width=1.0\linewidth]{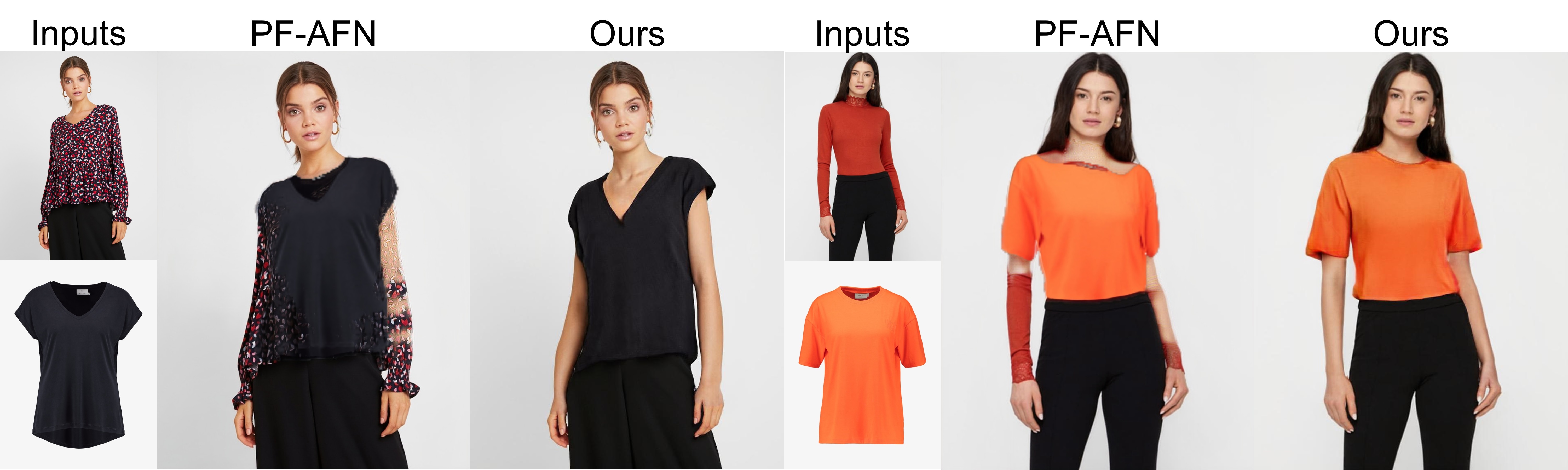}
    \caption{Qualitative comparison with PF-AFN on 1024×768 resolution.}
    \label{fig:PF}
\end{figure}
Following previous studies, we evaluate a paired setting and an unpaired setting, where the paired setting is to reconstruct the person image with the original clothing image, and the unpaired setting is to change the clothing item of the person image.
For paired setting, we evaluate our method using two widely-used metrics: Structural Similarity (SSIM)~\cite{wang2004image} and Learned Perceptual Image Patch Similarity (LPIPS)~\cite{zhang2018unreasonable}.
Additionally, to evaluate the unpaired setting, we measure Frechet Inception Distance (FID)~\cite{heusel2017gans} and Kernel Inception Distance (KID), which is a more descriptive metric than FID when the number of data is small.

\noindent \textbf{Ablation Study.}
Table~\ref{table:ablation} shows the effectiveness of the proposed feature fusion block and condition aligning.
Indeed, the benefits of fusion block and condition aligning are largely additive.
Notably, the model without feature fusion block and condition aligning yields suboptimal results, demonstrating the necessity of information exchange between the warping module and the segmentation map generator.

\noindent \textbf{Comparison with Baselines.} Table~\ref{table:main} demonstrates that our method outperforms the baselines for all evaluation metrics, especially at the 1024$\times$768 resolution.
The results indicate that CP-VTON and ACGPN can not handle the high-resolution images in the unpaired setting.
Furthermore, it is noteworthy that our framework surpasses VITON-HD, one of the state-of-the-art methods for high-resolution virtual try-on.
Although our try-on image generator is very similar to one of VITON-HD, our framework has superior performance due to the capability to produce high-quality conditions (\textit{i.e.}, segmentation map and warped clothing image).

\subsection{Comparison with Parser-free Virtual Try-on Methods}
Recently, several approaches~\cite{issenhuth2020not,ge2021parser} propose virtual try-on models that do not rely on a predicted segmentation map.
However, explicitly predicting a segmentation map helps the model distinguish the regions to be generated and the regions to be preserved, which is necessary for a high-resolution virtual try-on.
To verify this, we compare our model with PF-AFN~\cite{ge2021parser} on the high-resolution dataset.
Fig.~\ref{fig:PF} demonstrates that PF-AFN fails to remove the original clothing regions as it can not differentiate the parts to be generated and the parts to be left, resulting in significant artifacts in the outputs.
Moreover, Table~\ref{table:main} shows that our model outperforms PF-AFN by a large margin.
The results indicate that it is difficult to obtain convincing high-resolution results without predicting a segmentation map.

\section{Discussion}
\noindent\textbf{Limitation of Discriminator Rejection.}
The existing image-based virtual try-on approaches assume that test data is drawn from the same distribution as the training data.
However, in the real-world scenario, it is prevalent that the input images are taken at a different camera view from the training images or even do not contain humans.
Since the low-quality segmentation is often predicted due to such out-of-distribution inputs, our discriminator rejection is capable of filtering out the out-of-distribution inputs.
We believe that our discriminator rejection can be a solution to enhance the user experience in virtual try-on applications.

\section{Conclusion}
In this paper, we propose a novel architecture for high-resolution virtual, which performs warping clothes and segmentation generation simultaneously while exchanging information with each other.
The proposed try-on condition generator completely eliminates the misaligned region and solves the pixel-squeezing problem by handling the occlusion by body parts.
We also demonstrate that the discriminator of the condition generator can filter out the impaired segmentation results, which is practically helpful for real-world virtual try-on applications.
Extensive experiments show that our method outperforms the existing virtual try-on methods at 1024$\times$768 resolution.

\section*{Acknowledgement}
This work was supported by the Institute of Information \& communications Technology Planning \& Evaluation (IITP) grant funded by the Korea government(MSIT) (No. 2019-0-00075, Artificial Intelligence Graduate School Program(KAIST) and No.2021-0-02068, Artificial Intelligence Innovation Hub) and the National Research Foundation of Korea (NRF) grant funded by the Korean government (MSIT) (No. NRF-2022R1A2B5B02001913)


\clearpage
%
%
\bibliographystyle{splncs04}
\bibliography{reference}
\newpage
\appendix
\begin{center}
      {\bf APPENDIX}
    \end{center}

\section*{A. Implementation Details}

In this section, we describe the detailed architectures, hyper-parameters, and objective functions of the try-on condition generator and the image generator.

\begin{figure}[h!]
    \centering
    \includegraphics[width=1.0\linewidth]{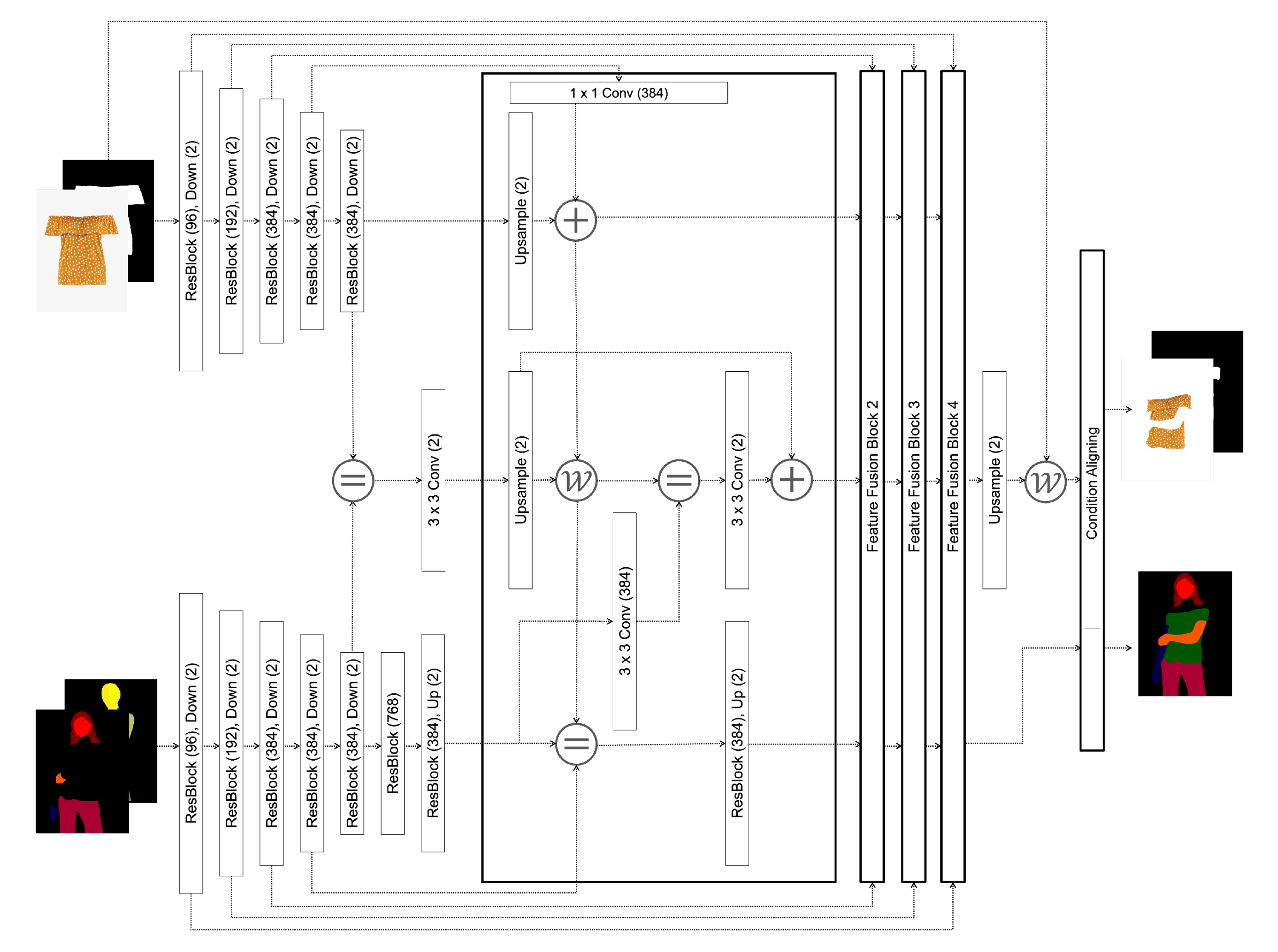}
    \vspace{-0.5cm}
    \caption{The detailed architecture of the try-on condition generator. (ResBlock ($n$), Up/Down ($f$)) denotes a residual block where the scaling factor is $f$ and the output channel is $n$. Conv ($m$) denotes a convolutional layer where the output channel is $m$.}
    \vspace{-0.5cm}
    \label{fig:architecture-detail}
\end{figure}

\noindent\textbf{Try-On Condition Generator.}
The try-on condition generator consists of two encoders and four feature fusion blocks, and each encoder is composed of five residual blocks.
The features of the last residual blocks are concatenated and passed to a 3$\times$3 convolutional layer, which generates the first flow map of the flow pathway.
Also, the last feature of the segmentation encoder is used as the input of the segmentation pathway (\textit{i.e.}, seg pathway) after passing through two residual blocks.
We employ two multi-scale discriminators for the conditional adversarial loss.
The visualization of the try-on condition generator architecture is in Fig.~\ref{fig:architecture-detail}.

During the training of our try-on condition generator, the model predicts $\hat I_c, \hat S_c$, and $\hat S$ at 256$\times$192 resolution. 
In the inference phase, before forwarding our try-on image generator, the segmentation map and the appearance flow obtained from the try-on condition generator are upscaled to 1024$\times$768. 
We down-sampled the inputs for the discriminator of our try-on condition generator by a factor of 2 to increase the receptive field.
In addition, we apply a dropout~\cite{srivastava2014dropout} to the discriminator to stabilize the training.
For hyper-parameters we used, $\lambda_{CE}, \lambda_{VGG}$, and $\lambda_{TV}$ are set to 10, 10, and 2, respectively.
The batch sizes for training our try-on condition generator and image generator are set to 8 and 4, respectively.
We train each module for 100,000 iterations.
The learning rates of the generator and the discriminator of the try-on condition generator are set to 0.0002.

\begin{figure}[h!]
    \centering
    \includegraphics[width=0.9\linewidth]{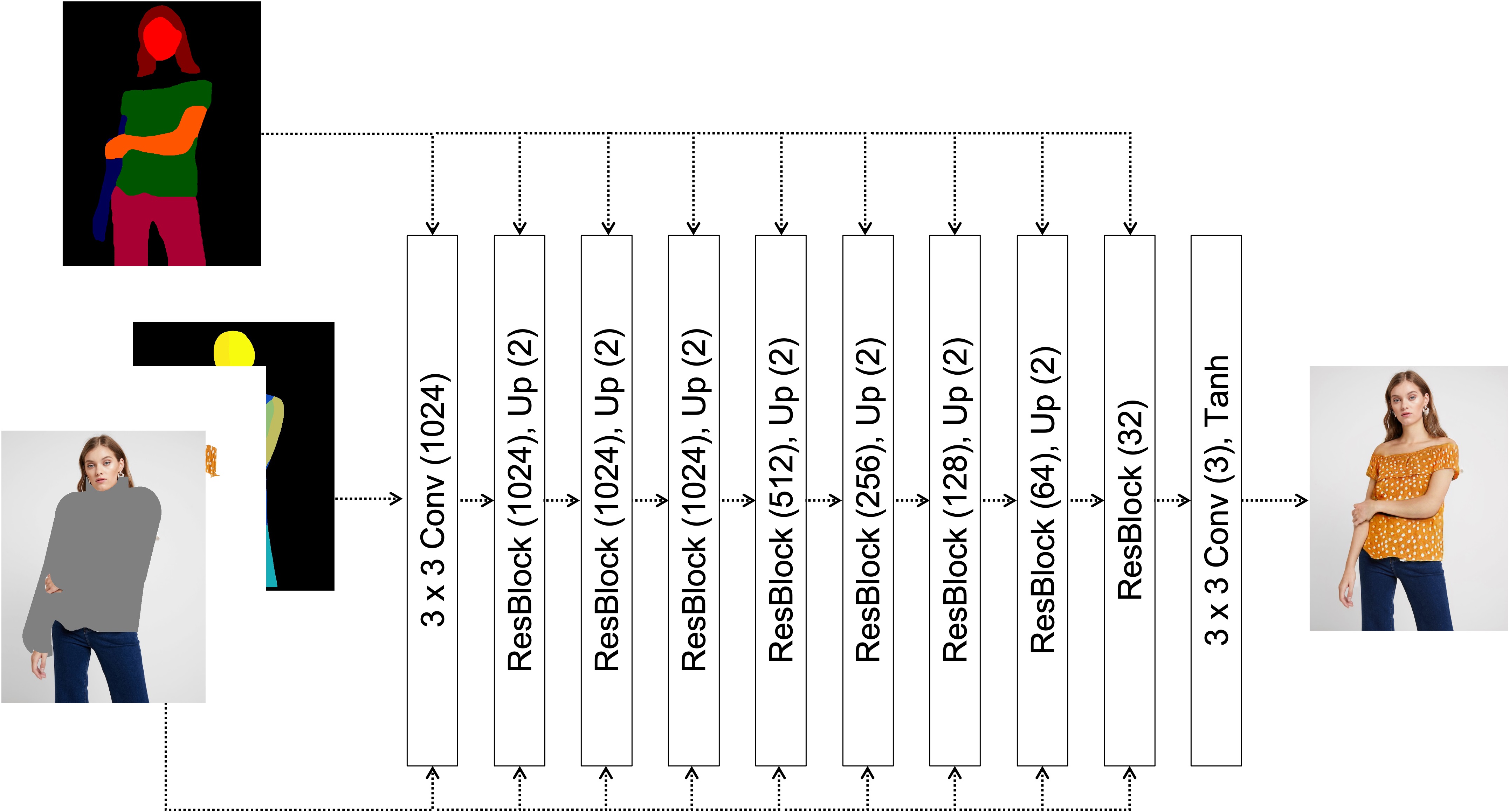}
    \caption{The detailed architecture of the try-on image generator. (ResBlock ($n$), Up ($f$)) denotes a residual block, where the scaling factor is $f$, and the output channel is $n$. Conv ($m$) denotes a convolutional layer where the output channel is $m$.}
    \vspace{-0.5cm}
    \label{fig:image-generator-detail}
\end{figure}

\noindent\textbf{Try-On Image Generator.}
We describe the detailed architecture of the try-on image generator as shown in Fig.~\ref{fig:image-generator-detail}.
The generator is composed of a series of residual blocks with upsampling layers, and two multi-scale discriminators are employed for the conditional adversarial loss.
Spectral normalization~\cite{miyato2018spectral} is applied to all the convolutional layers.

To train the try-on image generator, we utilize the same losses used in SPADE~\cite{park2019semantic} and pix2pixHD~\cite{wang2018high}.
Specifically, our full objective function consists of the conditional adversarial loss, the perceptual loss, and the feature matching loss.
Formally, our objective function is as follows:

\begin{equation}
    \mathcal{L}_{TOIG} = \mathcal{L}_{cGAN}^{TOIG} + \lambda_{VGG}^{TOIG}\mathcal{L}_{VGG}^{TOIG} + \lambda_{FM}^{TOIG}\mathcal{L}_{FM}^{TOIG},
\end{equation}
where $\mathcal{L}_{cGAN}^{TOIG}$, $\mathcal{L}_{VGG}^{TOIG}$, and $\mathcal{L}_{FM}^{TOIG}$ denote the conditional adversarial loss, the perceptual loss, and the feature matching loss~\cite{wang2018high}, respectively.
We use $\lambda_{VGG}^{TOIG}$ and $\lambda_{FM}^{TOIG}$ for hyper-parameters controlling relative importance between different losses.
For $\mathcal{L}_{GAN}^{TOIG}$, we employ the Hinge loss~\cite{lim2017geometric}. 
$\lambda_{VGG}^{TOIG}$ and $\lambda_{FM}^{TOIG}$ are set to 10.
The learning rates of the generator and the discriminator of the try-on image generator are set to 0.0001 and 0.0004, respectively.
We adopt the Adam optimizer with $\beta_{1}=0.5$ and $\beta_{2}=0.999$ for both modules.

\section*{B. Additional Experiments}

\begin{figure}[t!]
    \centering
    \includegraphics[width=1.0\linewidth]{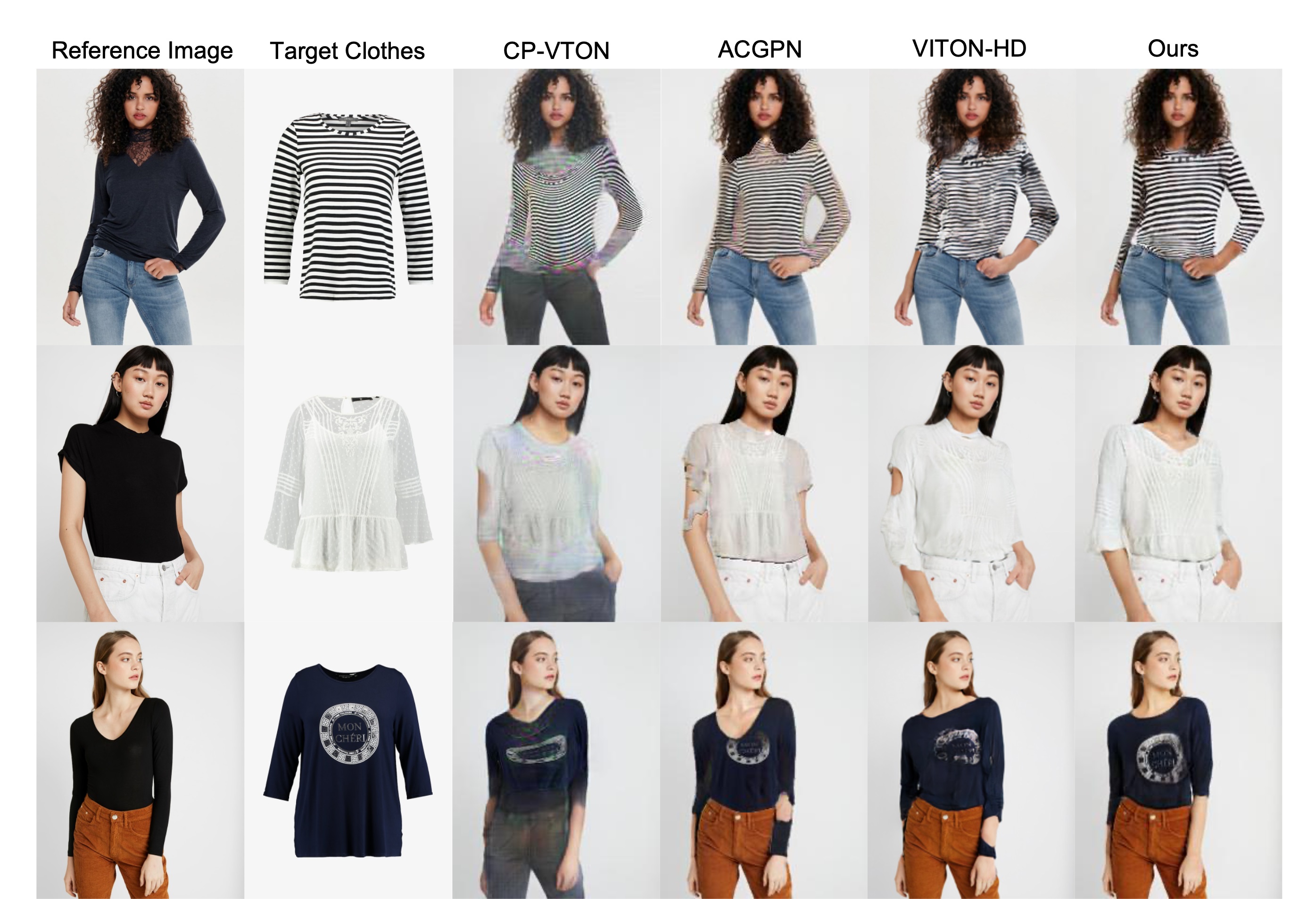}
    \caption{Qualitative comparison of the baselines (256$\times$192)}
    \label{fig:comparison-256}
\end{figure}

\begin{figure}
    \centering
    \includegraphics[width=1.0\linewidth]{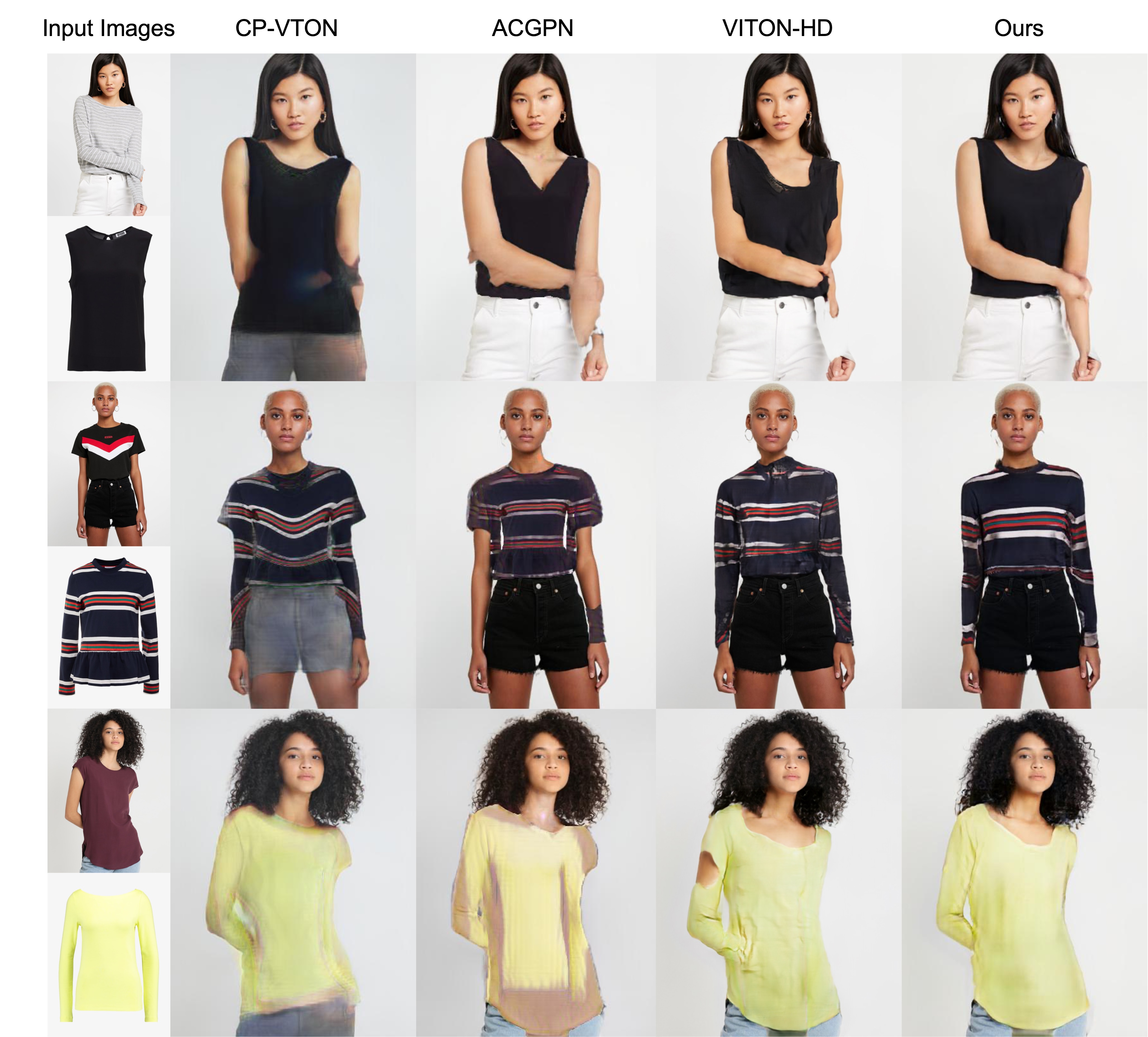}
    \caption{Qualitative comparison of the baselines (512$\times$384)}
    \label{fig:comparison-512}
\end{figure}

\begin{figure}[t!]
    \centering
    \includegraphics[width=1.0\linewidth]{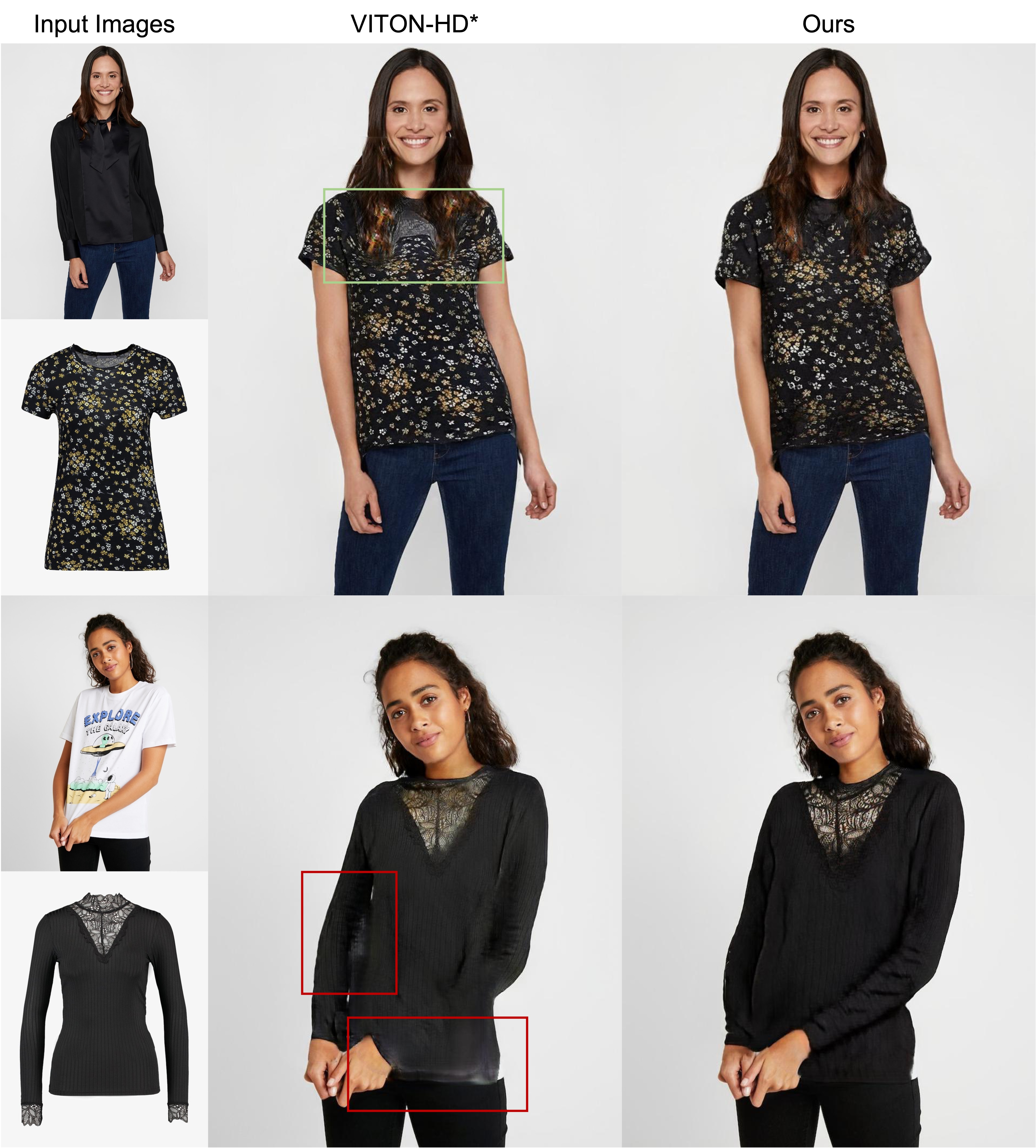}
    \caption{Qualitative comparison with VITON-HD* (1024$\times$768). VITON-HD* suffers from the misalignment and the pixel-squeezing artifacts indicated by green and red colored areas, respectively.}
    \label{fig:vitonhdclothflow}
\end{figure}

\noindent\textbf{Results on Different Resolutions.}
We provide the additional qualitative results for comparison across different resolutions (Fig.~\ref{fig:comparison-256}, and Fig.~\ref{fig:comparison-512}). 

\noindent\textbf{Comparison with the Variant of VITON-HD.}
Previous studies~\cite{han2019clothflow,chopra2021zflow} improve the performance of the geometric deformation for the target clothes by utilizing the appearance flow.
However, simply increasing the degree of freedom of the warping module cannot perfectly remove the artifacts caused by misalignment and pixel-squeezing.
To verify this, we further compare our method with VITON-HD*, the VITON-HD variant of which the clothes warping module is replaced by that of Clothflow~\cite{han2019clothflow}.
Since Clothflow is superior to the warping module of VITON-HD, VITON-HD* can reduce the misalignment region.

\begin{table}[t!]
\centering
\footnotesize
\begin{tabular}{l|cccc}
    \toprule
     & LPIPS$_{\downarrow}$ & SSIM$_{\uparrow}$  & FID$_{\downarrow}$ & KID$_{\downarrow}$ \\ 
    \midrule
    VITON-HD* & 0.070 & 0.875 & 11.55 & 0.2993    \\
    Ours & \textbf{0.065} & \textbf{0.892} & \textbf{10.91} & \textbf{0.1794} \\
    \bottomrule
\end{tabular}
\caption{Quantitative comparison with VITON-HD* at the 1024$\times$768 resolution. 
We describe the KID as a value multiplied by 100.}
\label{clothflow}
\end{table}

Despite the improvement of the warping module in VITON-HD, our model consistently outperforms the VITON-HD* in all evaluation metrics, as seen in Table~\ref{clothflow}.
Also, \textit{2nd} column in Fig.~\ref{fig:vitonhdclothflow} shows that VITON-HD* still suffers from the artifacts due to the misalignment.
Furthermore, increasing the degree of freedom of the warping module exacerbates the pixel-squeezing artifact, indicating that the use of appearance flow without proper occlusion handling can be harmful.
On the other hand, our model successfully solves both the misalignment and the pixel-squeezing problems, as shown in \textit{3rd} column in Fig.~\ref{fig:vitonhdclothflow}.

\begin{figure}[h!]
    \includegraphics[width=1.0\linewidth]{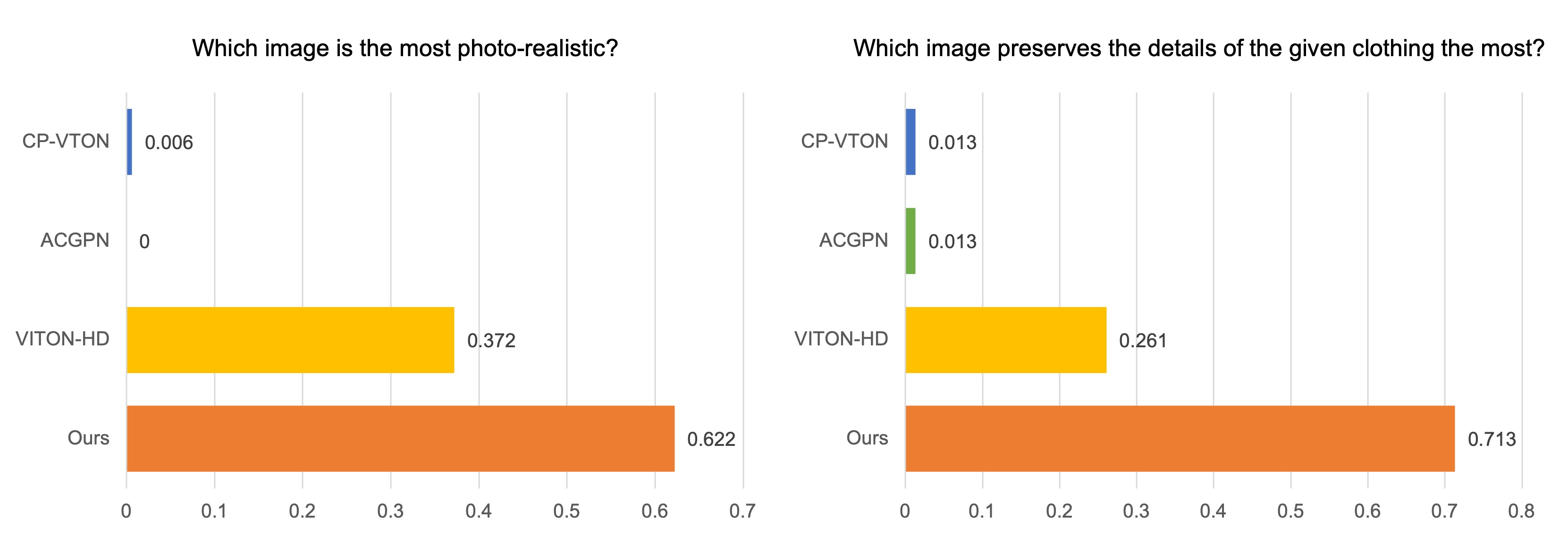}
    \vspace{-0.5cm}
    \caption{User study results.}
    \label{fig:userstudy}
    
    \includegraphics[width=1.0\linewidth]{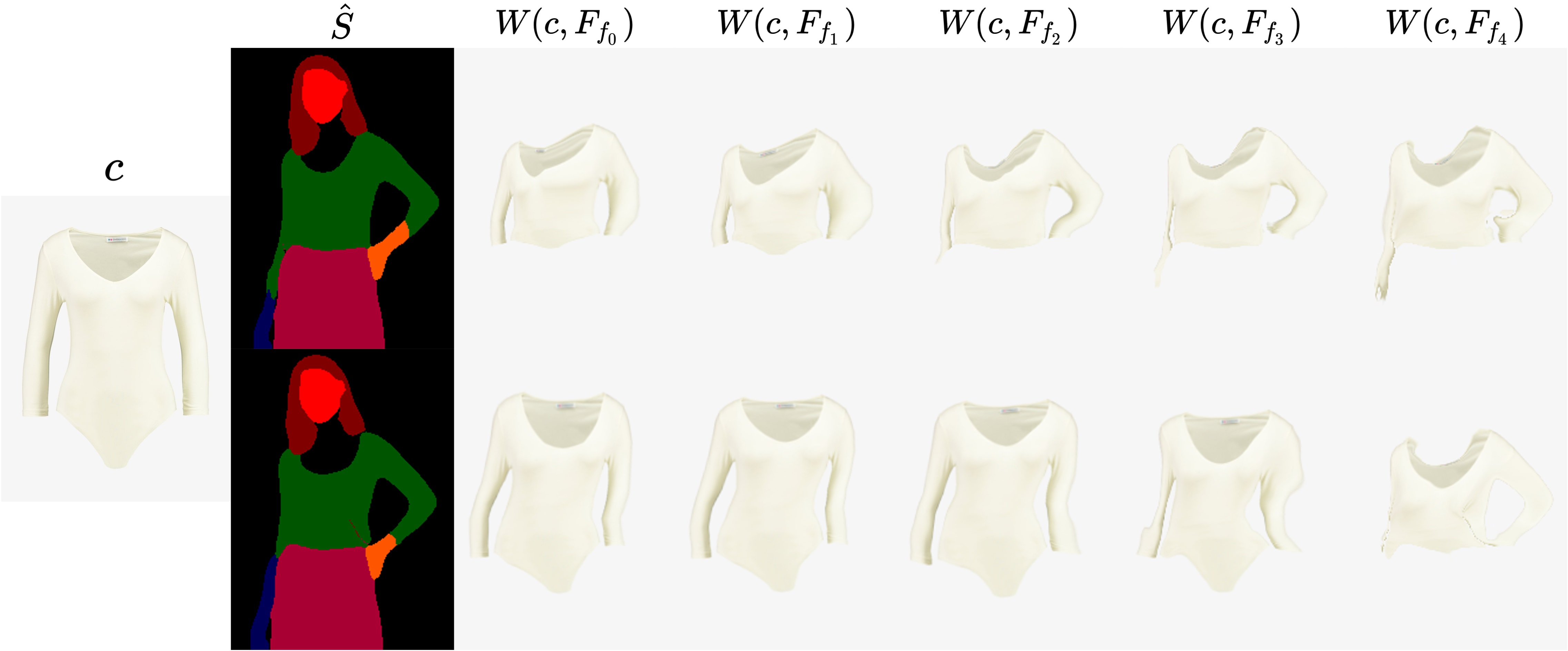}
    \vspace{-0.5cm}
    \caption{Effects of the multi-scale $L1$/VGG losses. $1st$ row: w/ multi-scale losses. $2nd$ row: w/o multi-scale losses.}
    \vspace{-0.5cm}
    \label{fig:multiscale}
\end{figure}

\noindent\textbf{User Study.}
We conduct a user study to further assess our model and other baselines at the 1024$\times$768 resolution.
Given the 30 sets of a reference image and a target garment image from the test set, the users are asked to choose an image among the synthesized results of our model and baselines according to the following questions: (1) Which image is the most photo-realistic? (2) Which image preserves the details of the given clothing the most?
In addition, a total of 21 participants participate in the user study.
Fig.~\ref{fig:userstudy} shows that our model achieves the highest average selection rate for both questions, indicating that our model synthesizes more perceptually convincing results and preserves the detail of the clothing items better than other baselines.

\noindent\textbf{Effectiveness of Multi-Scale L1/VGG Losses.}
During the training of the try-on condition generator, $\mathcal{L}_{L1}$ and $\mathcal{L}_{VGG}$ are directly applied to the intermediate flow estimations.
As shown in \textit{2nd} row of Fig.~\ref{fig:multiscale}, the model without the multi-scale losses has difficulty learning flow estimation in a coarse scale.
Multi-scale losses enable the model to learn the meaningful intermediate flow estimation, which is crucial for the coarse-to-fine generation of appearance flow.

\begin{figure}[b!]
    \centering
    \includegraphics[width=1.0\linewidth]{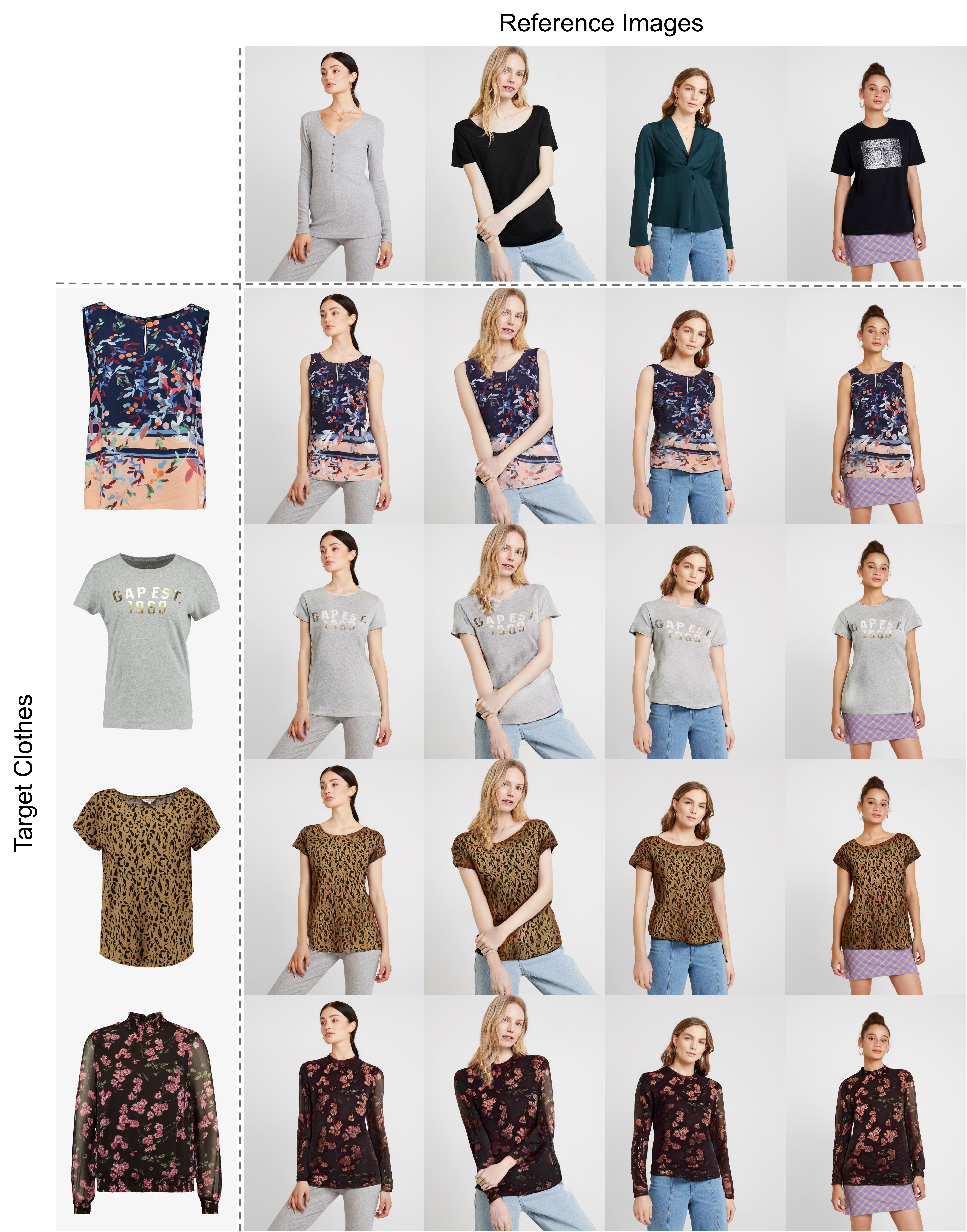}
    \caption{Qualitative results of our model (1024$\times$768).}
    \label{fig:results-matrix}
\end{figure}
\begin{figure}[h!]
    \centering
    \includegraphics[width=1.0\linewidth]{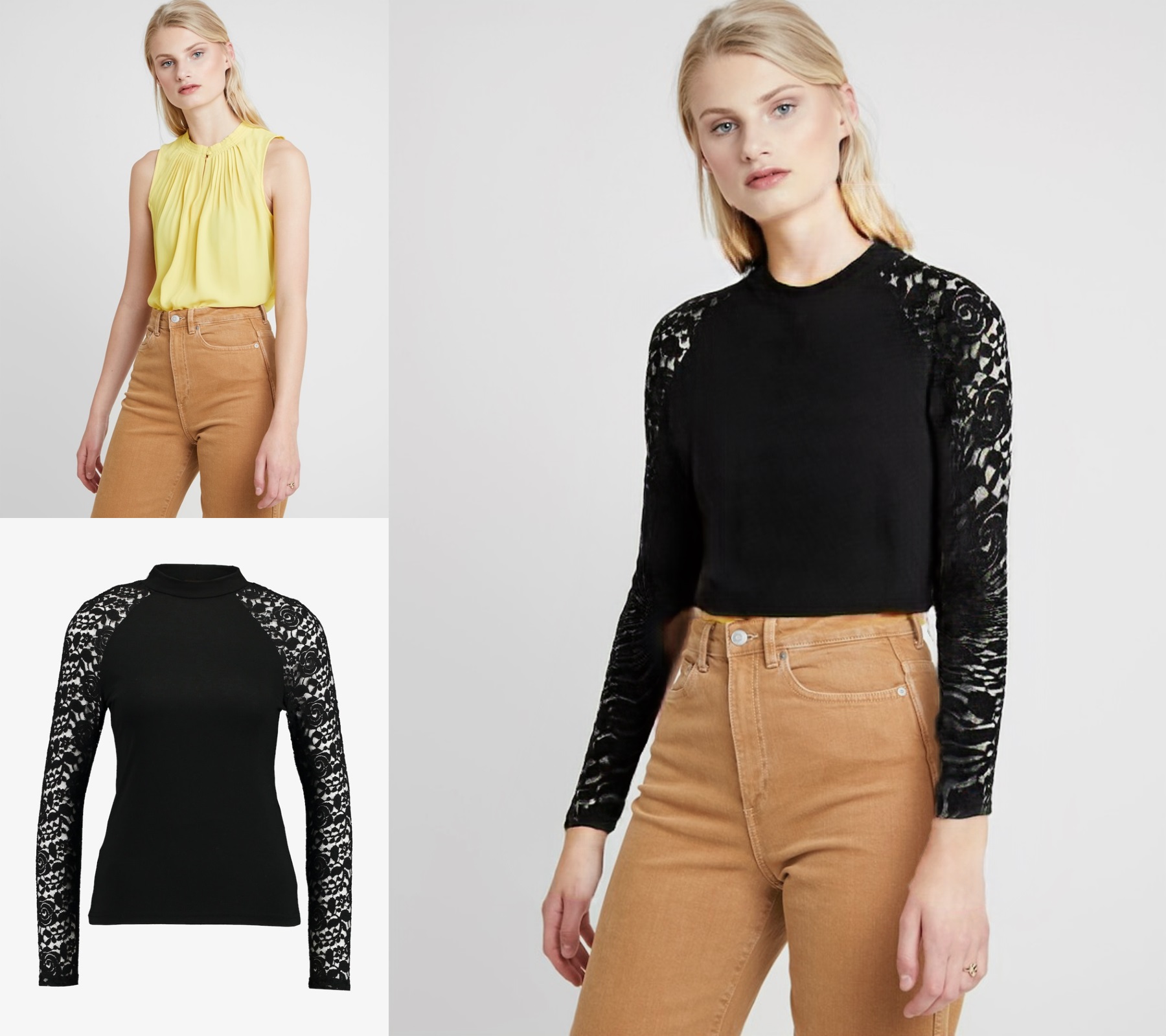}
    \caption{Qualitative results of our model (1024$\times$768). The reference image and the target clothes (\emph{left}), the synthesis image (\emph{right}).}
    \label{fig:results-big1}
\end{figure}
\begin{figure}[h!]
    \centering
    \includegraphics[width=1.0\linewidth]{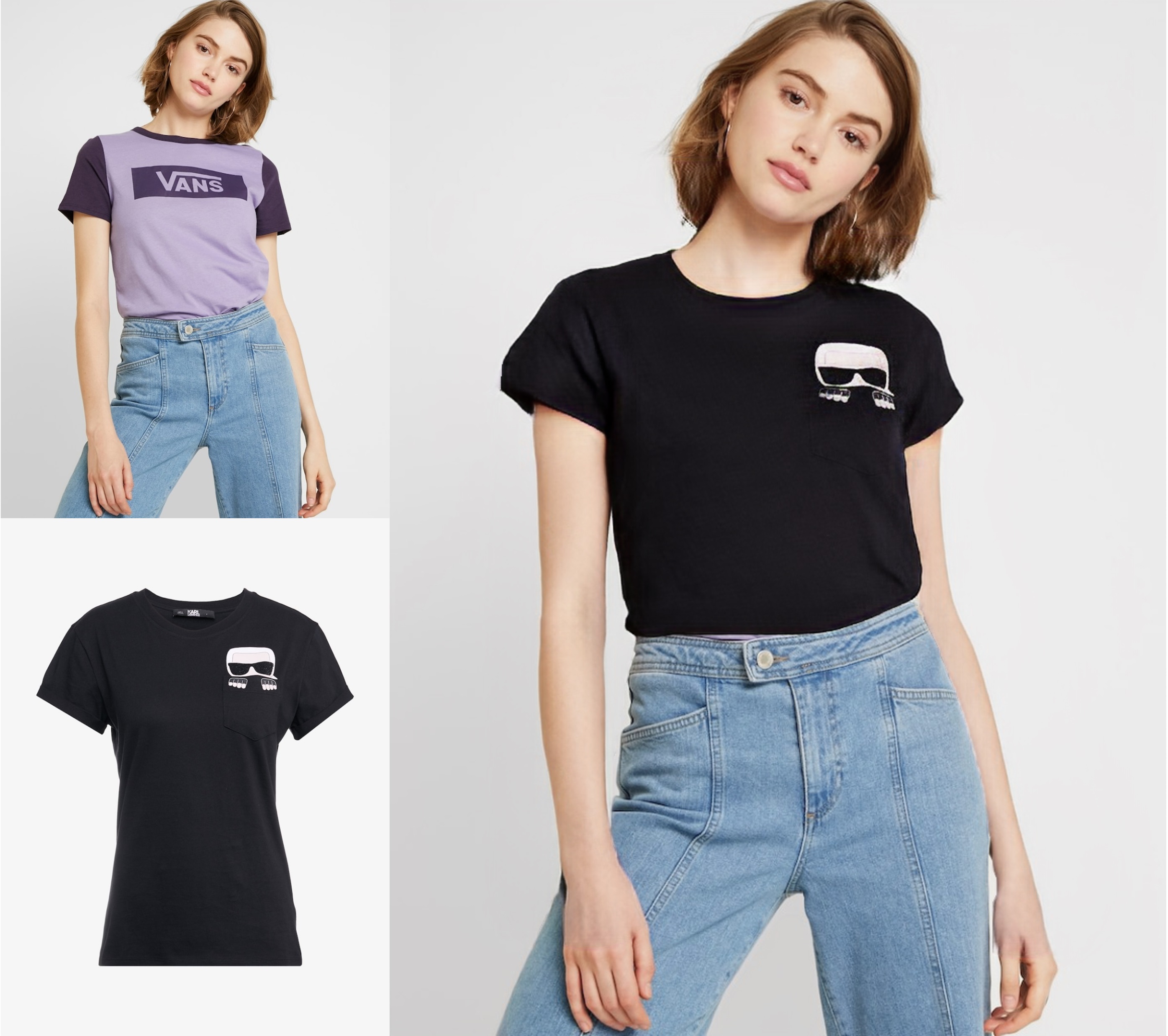}
    \caption{Qualitative results of our model (1024$\times$768). The reference image and the target clothes (\emph{left}), the synthesis image (\emph{right}).}
    \label{fig:results-big2}
\end{figure}
\begin{figure}[h!]
    \centering
    \includegraphics[width=1.0\linewidth]{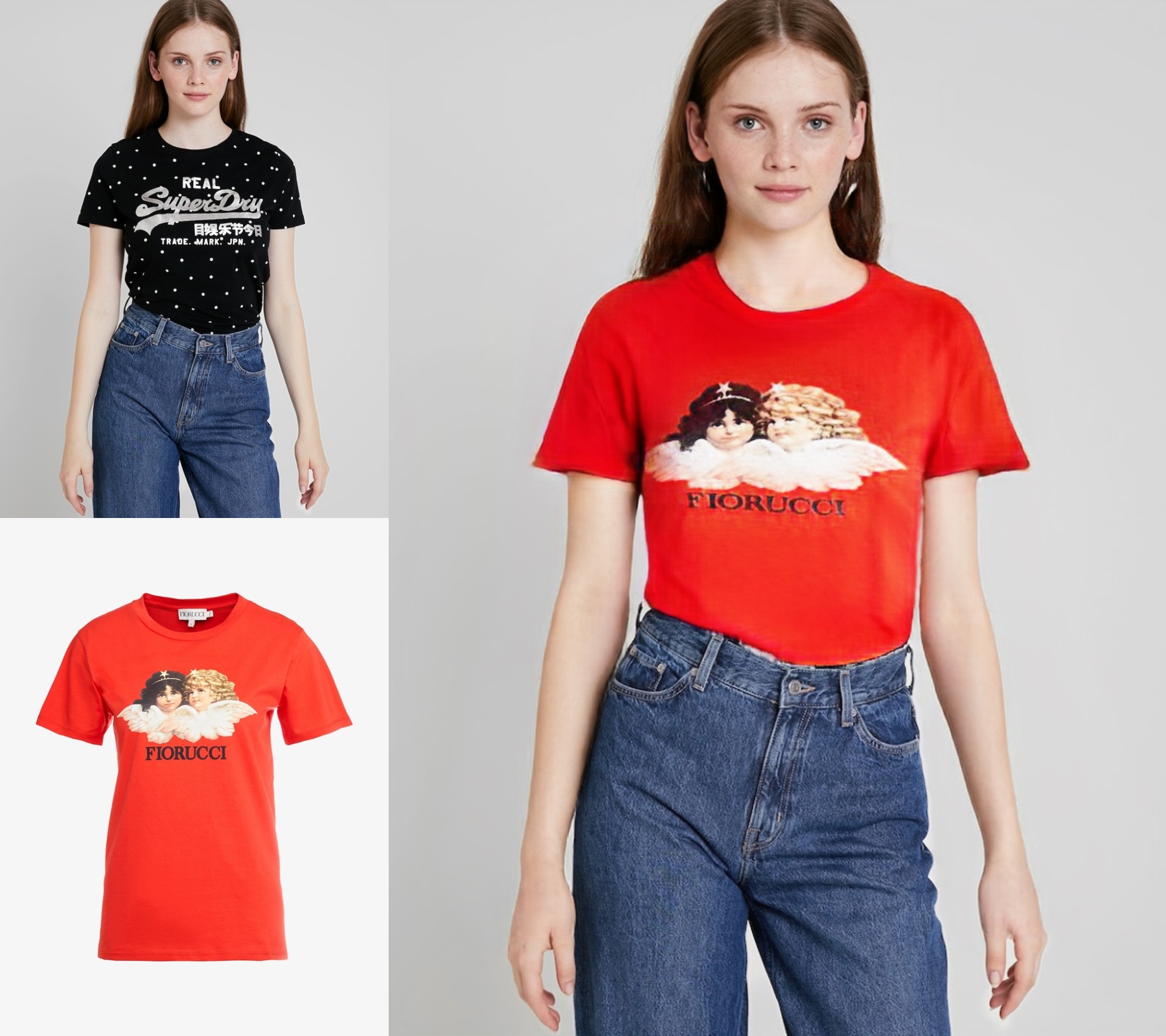}
    \caption{Qualitative results of our model (1024$\times$768). The reference image and the target clothes (\emph{left}), the synthesis image (\emph{right}).}
    \label{fig:results-big3}
\end{figure}

\noindent\textbf{Additional Results.}
We present additional qualitative results of our model.
Fig.~\ref{fig:results-matrix} shows the combination of different clothes and different people, and Fig.~\ref{fig:results-big1}-~\ref{fig:results-big3} shows the high-resolution synthesis results (\textit{i.e.}, 1024$\times$768).

\end{document}